\RequirePackage{fix-cm}
\documentclass{article}
\usepackage{graphicx}
\usepackage{textcomp}	
\usepackage{fancybox}
\usepackage{anyfontsize}	
\usepackage[utf8]{inputenc}	
\usepackage{tocloft} 
\usepackage{color}
\usepackage{soul}
\usepackage{natbib}
\usepackage{hyperref}
\usepackage{tablefootnote}
\usepackage{pdfpages}
\usepackage{pbox}
\usepackage[]{siunitx}
\usepackage{cleveref}
\usepackage{caption}
\usepackage{multirow}
\usepackage{float}
\usepackage{tabularx}
\newcolumntype{Y}{>{\centering\arraybackslash}X}
\makeatletter \let\cl@chapter\relax \makeatother 
\begin{document}

\title{Evaluation of contextual embeddings on less-resourced languages}
\author{Matej Ulčar$^1$,  Aleš Žagar$^1$, Carlos S.\ Armendariz$^2$, Andraž Repar$^3$ \\
Senja Pollak$^3$, Matthew Purver$^2$, Marko Robnik-Šikonja$^1$  \\
\ \\
$^1$ University of Ljubljana, Faculty of Computer and Information Science\\
Večna pot 113, 1000 Ljubljana, Slovenia\\
\texttt{\{matej.ulcar, ales.zagar, marko.robnik\}@fri.uni-lj.si} \\
\ \\
$^2$ Queen Mary University of London, Cognitive Science Research Group, \\
Mile End Road, London E1 4NS, United Kingdom \\
\texttt{\{c.santosarmendariz, m.purver\}@qmul.ac.uk} \\
\ \\
$^3$ Jožef Stefan Institute, \\
Jamova 39, 1000 Ljubljana, Slovenia \\
\texttt{\{andraz.repar, senja.pollak\}@ijs.si}
}

\date{}
\maketitle

\begin{abstract}
The current dominance of deep neural networks in natural language processing is based on contextual embeddings such as ELMo, BERT, and BERT derivatives. Most existing work focuses on English; in contrast, we present here the first multilingual empirical comparison of two ELMo and several monolingual and multilingual BERT models using 14 tasks in nine languages. In monolingual settings, our analysis shows that monolingual BERT models generally dominate, with a few exceptions such as the dependency parsing task, where they are not competitive with ELMo models trained on large corpora. In cross-lingual settings, BERT models trained on only a few languages mostly do best, closely followed by massively multilingual BERT models. 
\end{abstract} %

\section{Introduction}
Deep neural networks have dominated the area of natural language processing (NLP) for almost a decade. The establishment of contextual embeddings such as ELMo \citep{Peters2018} and BERT \citep{Devlin2019} have advanced many NLP tasks to previously unattainable performance, often achieving human levels. At the same time, newer generations of transformer-based \citep{Vaswani2017} neural language models have increased in size and training times to levels unachievable to anyone but large corporations. While training BERT with 110 million parameters is possible on 8 GPUs in about a month, the T5 model \citep{raffel2020exploring} contains 11 billion parameters and GPT-3 \citep{brown2020language} contains 175 billion parameters. These large models are trained on English. With the exception of Chinese, no similar-sized models exist for another language, while several BERT-sized models have sprouted for other languages. 

In this paper, we focus on empirical mono- and cross-lingual comparison of ELMo and BERT contextual models for less-resourced but technologically still relatively well-supported European languages. This choice stems from the enabling conditions for such a study: availability of contextual ELMo and BERT models and availability of evaluation datasets. These constraints and limited space have led us to select nine languages (Croatian, English\footnote{We included English for comparison with other languages and for cross-lingual knowledge transfer.}, Estonian, Finnish, Latvian, Lithuanian, Russian, Slovene, Swedish) and seven categories of datasets: named-entity recognition (NER), part-of-speech (POS) tagging, dependency  parsing (DP), analogies, contextual similarity (CoSimLex), terminology alignment, and the SuperGLUE suite of benchmarks (eight tasks). While some tasks are well-known and frequently used in the presentation of new models, others are used here for the first time in cross-lingual comparisons. We compare two types of ELMo models (described in \Cref{sec:ELMo}) and three categories of BERT models: monolingual, massively multilingual and moderately multilingual (trilingual to be precise). The latter models are specifically intended for cross-lingual transfer.

The aim of the study is  to compare i) the quality of different monolingual contextual models and ii) the success of cross-lingual transfer between similar languages and from English to less-resourced languages. While partial comparisons exist for individual languages (in particular English) and individual tasks, no systematic study has yet been conducted. This study fills this gap. 

The main contributions of the work are as follows.
\begin{enumerate}
\item The establishment of a set of mono- and cross-lingual datasets suitable for the evaluation of contextual embeddings in less-resourced languages.
\item The first systematic monolingual evaluation of ELMo and BERT contextual embeddings for a set of less-resourced languages. 
\item The first systematic evaluation of cross-lingual transfer using contextual ELMo and BERT models for a set of less-resourced languages. 
\end{enumerate}

The structure of the paper is as follows. \Cref{sec:related} outlines related work. In \Cref{sec:embeddings}, we describe the monolingual and cross-lingual embedding approaches used. We split them into four categories: baseline non-contextual fastText embeddings, contextual ELMo embeddings, cross-lingual maps for these, and BERT-based monolingual and cross-lingual models. In \Cref{sec:evaluation}, we present our evaluation scenarios, divided into settings and benchmarks. \Cref{sec:results} contains the results of the evaluations. We first cover the monolingual approaches, followed by the cross-lingual ones. We present our conclusions in \Cref{sec:conclusions}.

\section{Related works}
\label{sec:related}
Ever since their introduction, ELMo \citep{Peters2018} and BERT \citep{Devlin2019} have attracted  enormous attention from NLP researchers and practitioners. \citet{rogers2020primer} present a recent survey of over 150 papers investigating the information BERT contains, modifications to its training objectives and architecture, its overparameterization, and compression. 

At the time of its introduction, ELMo has been shown to outperform previous pretrained word embeddings like word2vec and GloVe on many NLP tasks \citep{Peters2018}. Later, BERT models turned out to be even more successful on these tasks \citep{Devlin2019} and many others \citep{wang2019superglue}. This fact would seemingly make ELMo obsolete. However, concerning the quality of extracted vectors, ELMo can be advantageous \citep{skvorc2020MICE}. Namely, the information it contains is condensed into only three layers, while multilingual BERT uses 14 layers and the useful information is spread across all of them \citep{tenney-etal-2019-bert}. For that reason, we empirically evaluate both ELMo and BERT.

There are two works analysing ELMo on several tasks in a systematic way. Both are limited to English. The original ELMo paper \citep{Peters2018} uses six tasks: question answering, named entity extraction, sentiment analysis, textual entailment, semantic role labelling, and coreference resolution. Introducing the GLUE benchmark, \citet{wang2019glue} analyzed ELMo on nine tasks: linguistic acceptability, movie review sentiment, paraphrasing, question answering (two tasks), text similarity, natural language inference (two tasks), and coreference resolution. 
For BERT, there are also two systematic empirical evaluations on English. The original BERT paper \citep{Devlin2019} used nine datasets in the GLUE benchmark and three more tasks: question answering, NER, and sentence completion. The SuperGLUE benchmark \citep{wang2019superglue} contains eight tasks where BERT was tested: four question answering tasks, two natural language inference tasks, coreference resolution, and word-sense disambiguation.

Other works study ELMo and/or BERT for individual tasks like NER \citep{Taille2020}, dependency parsing \citep{Li2019SelfattentiveBD}, diachronic changes \citep{rodina2020elmo},  sentiment analysis \citep{robnik2021slovenscina20}, or coreference resolution \citep{joshi2019bert}. Several papers introduce language specific ELMo or BERT models and evaluate it on tasks available in that language, e.g., Russian \citep{kuratov2019adaptation}, French \citep{martin-etal-2020-camembert}, or German \citep{risch2019hpidedis}. The instances of these models used in our evaluation are described in \Cref{sec:monoBERTs}. 

In cross-lingual settings, most works compare massively multilingual BERT models such as mBERT \citep{Devlin2019} and XLM-R \citep{conneau2019unsupervised}. \citet{ulcar2020xlbert} trained several  trilingual BERT models suitable for transfer from English to less-resourced similar languages. The massively multilingual and trilingual models are described in \Cref{sec:multiBERTs} and \Cref{sec:ourBERT}, respectively. 

In contrast to the above studies, our analysis of ELMo and BERT includes nine languages (including English), 14 tasks, and covers both monolingual and cross-lingual evaluation. By translating SuperGLUE tasks into Slovene, we offer the first cross-lingual evaluation of multilingual BERT models in this benchmark suite and also compare human and machine translations.

\section{Cross-lingual and contextual embedding}
\label{sec:embeddings}
In this section, we briefly describe the monolingual and cross-lingual approaches that we compare.
In \Cref{sec:baseline}, we briefly present the non-contextual fastText baseline, and in \Cref{sec:ELMo}, the contextual ELMo embeddings. Mapping methods for the explicit embedding spaces produced by these two types of approaches are discussed in \Cref{sec:XLmaps}. We describe large pretrained language models  based on the transformer neural network architecture (i.e.\ BERT variants) in \Cref{sec:BERT}. 

\subsection{Baseline fastText embeddings}
\label{sec:baseline}
\label{sec:fasttext}

 As deep neural networks became the predominant learning method for text analytics, it was natural that they also gradually became the method of choice for text embeddings. A procedure common to these embeddings is to train a neural network on one or more semantic text classification tasks and then take the weights of the trained neural network as a representation for each text unit (word, n-gram, sentence, or document). 
The labels required for training such a classifier come from huge corpora of available texts. Typically, they reflect word co-occurrence, like predicting the next or previous word in a sequence or filling in missing words but may be extended with other related tasks, such as sentence entailment. The positive instances for the training are obtained from texts in the used corpora, while the negative instances are mainly obtained with negative sampling (sampling from instances that are highly unlikely related).
 
 \citet{mikolov2013distributed} introduced the word2vec method and trained it on a huge Google News data set (about 100 billion words).  The pretrained  300-dimensional vectors for 3 million English words and phrases are publicly available\footnote{\url{https://code.google.com/archive/p/word2vec/}}. 
 Word2vec consists of two related methods, \emph{continuous bag of words (CBOW)} and \emph{skip-gram}.
 Both methods construct a neural network to classify co-occurring words by taking as an input a word and its $d$ preceding and succeeding words, e.g., $\pm ~ 5$ words. 
 
\citet{Bojanowski2017} developed the fastText method, built upon the word2vec method but introduced subword information, which is more appropriate for morphologically rich languages such as the ones processed in this work. They took the skip-gram method from word2vec and edited the scoring function used to calculate the probabilities. In the word2vec method, this scoring function is equal to a dot product between two word vectors. For words $w_t$ and $w_c$ and their respective vectors $u_t$ and $u_c$, the scoring function $s$ is equal to $s(w_t, w_c) = \mathbf{u}_t^\top\mathbf{u}_c$. The scoring function in fastText is a sum of dot products for each subword (i.e. character n-gram) that appears in the word $w_t$:
$$
s(w_t, w_c) = \sum_{g\in G_t}\mathbf{z}_g^\top\mathbf{u_c},
$$
where $\mathbf{z}_g$ is a vector representation of an n-gram (subword) $g$ and $G_t$ is a set of all n-grams (subwords) appearing in $w_t$. As fastText is conceptually very similar to word2vec, we do not treat them as different methods but only test fastText as the baseline.

\subsection{ELMo embeddings}
\label{sec:ELMo}
ELMo (Embeddings from Language Models) embedding \citep{Peters2018} is an example of a pretrained transfer learning model. 
The first layer is a CNN (Convolutional Neural Network) layer, which operates on a character level. This layer is context-independent, so each word always gets the same embedding, regardless of its context. It is followed by two biLM (bidirectional language model) layers. A biLM layer consists of two concatenated LSTMs \citep{hochreiter1997-lstm}. The first LSTM predicts the following word, based on the given past words, where each word is represented by the embeddings from the CNN layer. The second LSTM predicts the preceding word based on the given following words. The second LSTM layer is equivalent to the first LSTM, just reading the text in reverse.

The actual embeddings are constructed from the internal states of a bidirectional LSTM neural network. Higher-level layers capture context-dependent aspects, while lower-level layers capture aspects of syntax \citep{Peters2018}. 
To train the ELMo network, one inputs one sentence at a time. The representation of each word depends on the whole sentence, i.e. it reflects the contextual features of the input text and thereby polysemy of words. For an explicit word representation, one can use only the top layer. Still, more frequently, one combines all layers into a vector.
The representation of a word or a token $t_k$ at position $k$ is composed of
\begin{equation}
R_k  =  \{ x_k^{LM}, \overrightarrow{h}_{k,j}^{LM}, \overleftarrow{h}_{k,j}^{LM} ~|~ j=1,\dots,L\}  
\end{equation}
where $L$ is the number of layers (ELMo uses $L=2$), index $j$ refers to the level of bidirectional LSTM network, $x$ is the initial token representation (either word or character embedding), and $h^{LM}$ denotes hidden layers of forward or backward language model. 

In NLP tasks, any set of these embeddings may be used; however, a weighted average is usually used. The weights of the average are learned during the training of the model for the specific task. Additionally, an entire ELMo model can be fine-tuned on a specific end task.


We compare two variants of ELMo models. The ELMoForManyLangs pro\-ject (EFML) \citep{che-EtAl:2018:K18-2}  trained ELMo models for several languages but used relatively small datasets of 20 million words randomly sampled from the raw text released by the CONLL 2017 shared task (wikidump + common crawl) \citep{Ginter2017}. 
We group ELMo models using the same architecture, but trained on much larger datasets (from 270 million to 5,5 billion words) under the name L-ELMo (Large ELMo). In L-ELMo we include the original English 5.5B ELMo model\footnote{\url{https://allennlp.org/elmo}}, Russian ELMo model trained by DeepPavlov\footnote{\url{https://github.com/deepmipt/DeepPavlov}} on the Russian WMT News, and the ELMo models trained by the EMBEDDIA project\footnote{\href{http://hdl.handle.net/11356/1277}{http://hdl.handle.net/11356/1277}} for seven languages (Croatian, Estonian, Finnish, Latvian, Lithuanian, Slovene, and Swedish) \citep{ulcar2020high}. 

\subsection{Cross-lingual maps for fastText and ELMo}
\label{sec:XLmaps}

Cross-lingual alignment methods take precomputed word embeddings for each language and align them with the optional use of bilingual dictionaries. The goal of alignments is that the embeddings for words with the same meaning shall be as close as possible in the final vector space. \citet{Sogaard2019} overview the area of cross-lingual embeddings. A comprehensive summary of existing mapping approaches can be found in \citep{artetxe2018generalizing}. Special cross-lingual mapping techniques for contextual ELMo embeddings are presented in \citep{ulcar2021GAN}.

Context-dependent embedding models calculate a word embedding for each word's occurrence; thus, a word gets a different vector for each context. Mapping such vector spaces from different languages is not straightforward. \citet{Schuster2019} observed that vectors representing different occurrences of each word form clusters. They averaged the vectors for each word occurrence so that each word was represented with only one vector, a so-called anchor. They applied the same procedure to both languages and aligned the anchors using the supervised or unsupervised method of MUSE \citep{Conneau2018}. However, this method comes with a loss of information. Many words have multiple meanings, which cannot be averaged. For example, the word \emph{mouse} can mean a small rodent or a computer input device. Context-dependent models correctly assign significantly different vectors to these two meanings since they appear in different contexts. Further, a word in one language can be represented with several different words (one for each meaning) in another language or vice versa. By averaging the contextual embedding vectors, we lose these distinctions in meaning. 

To align contextual ELMo embeddings, \citet{ulcar2021GAN} developed four methods that take different contexts and word meanings into account. All methods require a different type of contextual mapping datasets, described below. The datasets map not only words but also their contexts. Two mappings follow the existing mapping techniques used for static embeddings and assume that the mapping spaces are isomorphic, while the other two drop this assumption. The isomorphic mapping methods are called Vecmap \citep{artetxe2018generalizing} and MUSE \citep{Conneau2018}, and two non-isomorphic mapping methods are named ELMoGAN-O and ELMoGAN-10k \citep{ulcar2021GAN}. 

To form a cross-lingual mapping between contextual embeddings, a word in one language has to be represented with several different words (one for each meaning) in another language. 
For that, we require two resources: a sentence aligned parallel corpus of the two covered languages and their bilingual dictionary. The dictionary alone is not sufficient, as the words are not given in the context. Therefore, we can not use it for the alignment of contextual embeddings.
The parallel corpus alone is also not sufficient as the alignment is on the level of paragraphs or sentences and not on the level of words. By combining both resources, we take a translation pair from the dictionary and find sentences in the parallel corpus, with one word from the pair present in the sentence of the first language and the second word from the translation pair present in the second language sentence. As a result, we get matching words in matching contexts (sentences) which can serve as contextual anchor points. The details are presented in \citep{ulcar2021GAN}.

\subsection{BERT embeddings}
\label{sec:BERT}

\label{sec:bert}
BERT (Bidirectional Encoder Representations from Transformers) embedding \citep{Devlin2019} generalises the idea of language models (LM) to masked language models (MLM). The MLM randomly masks some of the tokens from the input, and the task of LM is to predict the missing token based on its neighbourhood. BERT  uses the transformer architecture of neural networks \citep{Vaswani2017} in a bidirectional sense and further introduces the task of predicting whether two sentences appear in a sequence. 
The input representation of BERT are sequences of tokens representing subword units. The input to the BERT encoder is constructed by summing the embeddings of corresponding tokens, segments, and positions. Some widespread words are kept as single tokens; others are split into subwords (e.g., frequent stems, prefixes, suffixes---if needed, down to single letter tokens). The original BERT project offers pretrained English, Chinese, Spanish, and multilingual models.

To use BERT in classification tasks only requires adding connections between its last hidden layer and new neurons corresponding to the number of classes in the intended task. Then, the fine-tuning process is applied to the whole network; all the parameters of BERT and new class-specific weights are fine-tuned jointly to maximise the log-probability of the correct labels.

BERT has shown excellent performance on 11 NLP tasks:  8 from GLUE language understanding benchmark \citep{wang2019glue}, question answering, named entity recognition, and common-sense inference \citep{Devlin2019}. The performance on monolingual tasks has often improved upon ELMo. However, while multilingual BERT covers 104 languages, its subword dictionary comprises tokens from all covered languages, which might not be optimal for a particular language. Further, similarly to ELMo, its training and tuning are computationally highly demanding tasks out of reach for most researchers. 

Below we first describe the massively multilingual models, followed by monolingual and trilingual models used in our experiments.

\subsubsection{Massively multilingual BERT and RoBERTa models}
\label{sec:multiBERTs}
The multilingual BERT model (mBERT) \citep{Devlin2019} was trained simultaneously on 104 languages, using the available Wikipedia texts in these languages. The mBERT model provides a representation in which the languages are embedded in the same space without requiring further explicit cross-lingual mapping. This massively multilingual representation might be sub-optimal for any specific language or a subset of languages. 

Deriving from BERT, \citet{liu2019roberta} developed RoBERTa, which drops the sentence inference training task (predicting if two given sentences are consecutive or not) and keeps only masked token prediction. Unlike BERT, which generates masked corpus as a training dataset in advance, RoBERTa randomly masks a given percentage of tokens on the fly. In that way, in each epoch, a different subset of tokens get masked. \citet{conneau2019unsupervised} used RoBERTa architecture to train the massive multilingual XLM-RoBERTa (XLM-R) model, using 100 languages, akin to the mBERT model.

\subsubsection{Monolingual BERT-like models}
\label{sec:monoBERTs}
Following the success of BERT, similar large pretrained transformer language models appeared in other languages. 

\citet{kuratov2019adaptation} trained a monolingual Russian BERT (RuBERT) on Russian Wikipedia and news corpus. They used multilingual BERT (mBERT) to initialise all the model weights, except for the first layer embeddings, where they replaced the mBERT's vocabulary with Russian-only vocabulary. They offer the model via open source DeepPavlov library\footnote{\href{https://github.com/deepmipt/DeepPavlov}{https://github.com/deepmipt/DeepPavlov}}. 

Monolingual Finnish BERT (FinBERT) \citep{virtanen2019multilingual} model was trained on a 3.3 billion token corpus, composed of news (Yle, STT), online discussions (Suomi24) and internet crawl of Finnish websites. The online discussions part of the corpus represents more than half of the entire training data. FinBERT model shares the architecture with the BERT-base model, with 12 transformer layers and the hidden layer size of 768.

Estonian (EstBERT) \citep{tanvir2020estbert}, Latvian (LVBERT) \citep{znotins2020lvbert}, and Swedish (KB-BERT) \citep{swedish-bert} BERT models were all trained in the same manner as FinBERT. Estonian EstBERT was trained on a 1.1 billion word Estonian National Corpus 2017, comprised of Estonian Reference Corpus (90s–2008), Estonian Web (2013 and 2017), and Estonian Wikipedia (2017). Latvian LVBERT was trained on a relatively small corpus with 500 million tokens. It consists mostly of articles and comments from various news portals, while including also Latvian Balanced corpus LVK2018 and Latvian Wikipedia. 
National Library of Sweden (KB) trained KB-BERT on modern Swedish corpora, using resources from 1940 to 2019. The 3.5 billion word corpora are composed mostly of digitised newspapers and include also government publications, Swedish Wikipedia, comments from online forums, etc.
The quality of these monolingual BERT models varies, mostly depending on the training datasets size and quality. 


Slovene (SloBERTa) and Estonian (Est-RoBERTa) monolingual models were trained on large non-public high-quality datasets within the EMBEDDIA project\footnote{\href{http://www.embeddia.eu}{http://www.embeddia.eu}}. Both models closely follow the architecture and training approach of the Camembert base model \citep{martin-etal-2020-camembert}, which is itself based on RoBERTa. Both models have 12 transformer layers and approximately 110 million parameters. SloBERTa was trained for 200,000 steps (about 98 epochs) on Slovene corpora, containing 3.47 billion tokens in total. The corpora are composed of general language corpus, web-crawled texts, academic writings (BSc/BA, MSc/MA and PhD theses) and texts from Slovenian parliament. Est-RoBERTa was trained for about 40 epochs on Estonian corpora, containing mostly news articles from Ekspress Meedia, in total 2.51 billion tokens. 
The subword vocabularies contain 32,000 tokens for SloBERTa model and 40,000 tokens for Est-RoBERTa model. Both models are publicly available via the popular Hugging Face library\footnote{\href{https://huggingface.co/EMBEDDIA/sloberta}{https://huggingface.co/EMBEDDIA/sloberta}}\textsuperscript{,}\footnote{\href{https://huggingface.co/EMBEDDIA/est-roberta}{https://huggingface.co/EMBEDDIA/est-roberta}} and for individual download from CLARIN\footnote{\href{http://hdl.handle.net/11356/1397}{http://hdl.handle.net/11356/1397}}\textsuperscript{,}\footnote{\href{https://doi.org/10.15155/9-00-0000-0000-0000-00226L}{https://doi.org/10.15155/9-00-0000-0000-0000-00226L}}.

BERTić \citep{ljubesic-lauc-2021-bertic} is a transformer-based pretrained model using the Electra approach  \citep{clark2019electra}. Electra models train a smaller generator model and the main, larger discriminator model whose task is to discriminate whether a specific word is an original word from the text or a word generated by the generator model. The authors claim that the Electra approach is computationally more efficient than the BERT models based on masked language modelling. BERTić is a BERT-base sized model (110 million parameters and 12 transformer layers), trained on crawled texts from the Croatian, Bosnian, Serbian and Montenegrin web domains. While BERTić is a multilingual model, we use it as a monolingual model and apply it to the Croatian language datasets. Two reasons are supporting this decision. First, most training texts are  Croatian (5.5 billion words out of 8 billion). Second, the covered South Slavic languages are closely related, mutually intelligible, and are classified under the same HBS (Serbo-Croatian) macro-language by the ISO-693-3 standard.

\subsubsection{Trilingual BERT models}
\label{sec:ourBERT}
While massively multilingual models allow for a good cross-lingual transfer of trained models, they contain a relatively small input dictionary for each language and most of the words are composed of several tokens. A possible solution is to build BERT models on fewer similar languages. \citet{ulcar2020xlbert} constructed trilingual models featuring two similar languages and one highly resourced language (English). Because these models are trained on a small number of languages, they better capture each of them and offer better monolingual performance. At the same time, they can be used in a cross-lingual manner for knowledge transfer from a high-resource language to a low-resource language or between similar languages.

We analyze three trilingual models, the first trained on Slovene, Croatian and English data (CroSloEngual BERT), the second  on Estonian, Finnish and English (FinEst BERT), and the third on Latvian, Lithuanian and English (LitLat BERT). The models are publicly available via the popular Huggingface library\footnote{\href{https://huggingface.co/EMBEDDIA/crosloengual-bert}{https://huggingface.co/EMBEDDIA/crosloengual-bert}}\textsuperscript{,}\footnote{\href{https://huggingface.co/EMBEDDIA/finest-bert}{https://huggingface.co/EMBEDDIA/finest-bert}}\textsuperscript{,}\footnote{\href{https://huggingface.co/EMBEDDIA/litlat-bert}{https://huggingface.co/EMBEDDIA/litlat-bert}} and for individual download from CLARIN\footnote{\href{http://hdl.handle.net/11356/1317}{http://hdl.handle.net/11356/1317}}\textsuperscript{,}\footnote{\href{https://doi.org/10.15155/9-00-0000-0000-0000-0021CL}{https://doi.org/10.15155/9-00-0000-0000-0000-0021CL}}\textsuperscript{,}\footnote{\href{http://hdl.handle.net/20.500.11821/42}{http://hdl.handle.net/20.500.11821/42}}.
Each model was trained on deduplicated corpora from all three languages. 

FinEst BERT and CroSloEngual BERT were trained on BERT-base architecture \citep{ulcar2020xlbert}, using bert-vocab-builder\footnote{\href{https://github.com/kwonmha/bert-vocab-builder}{https://github.com/kwonmha/bert-vocab-builder}} to produce wordpiece vocabularies (composed of subword tokens) from the given corpora. The created wordpiece vocabularies contain 74,986 tokens for FinEst and 49,601 tokens for the CroSloEngual model.
The training dataset is a masked corpus. The training randomly masked 15\% of the tokens in the corpus and repeated the process five times, each time masking different 15\% of the tokens. The dataset is thus five times larger than the original corpora.
On this data, trilingual BERT models were trained for about 40 epochs, which is approximately the same as multilingual BERT.

Later LitLat BERT is based on the RoBERTa architecture, which has proven more robust and better performing than BERT. It offers two practical benefits over the original BERT approach. By dropping the next-sentence prediction training task, corpora shuffled on the sentence level can be used in training at the expense of more limited context (compared to the original 512 tokens used in BERT). 
The second benefit is that it allows for training on multiple GPUs out of the box, while BERT can only be trained on a single GPU unless complex workarounds are implemented. 
In training, the Lithuanian, Latvian and English corpora were split into three sets, train, eval and test. Train dataset contains $99\%$ of all the corpora; the other two sets contain $0.5\%$ each. The sentencepiece algorithm\footnote{\href{https://github.com/google/sentencepiece}{https://github.com/google/sentencepiece}} produced subword byte-pair-encodings (BPE) from a given train dataset. The created subword vocabulary contained 84,200 tokens. The model was trained for 40 epochs, with a maximum sequence length of 512 tokens. Like with FinEst BERT and CroSloEngual BERT, 15\% of the tokens were randomly masked during the training.

\section{Evaluation scenarios}
\label{sec:evaluation}
In this section, we describe the evaluation scenarios. First, in \Cref{sec:datasets}, we describe the datasets and evaluation metrics used in the evaluation tasks. In \Cref{sec:evalSettings}, we describe the settings of monolingual and cross-lingual evaluation experiments.

\subsection{Datasets and evaluation metrics}
\label{sec:datasets}
We used six categories of datasets: NER, POS-tagging, dependency parsing,  analogies, CoSimLex, and SuperGLUE.  Each category contains datasets from several languages, and some contain several types of tasks (e.g., SuperGLUE). The categories are shortly described below.

\subsubsection{Named entity recognition}
In the NER experiments, we use datasets in nine languages: Croatian, English, Estonian, Finnish, Latvian, Lithuanian, Russian, Slovene and Swedish.
The number of sentences and tags present in the datasets is shown in Table~\ref{tab:ner-datasets}.
The label sets used in datasets for different languages vary, meaning that some contain more fine-grained labels than others.
To make results across different languages consistent, we trim labels in all datasets to the four common ones: location (LOC), organisation (ORG), person (PER), and ``no entity'' (OTHR). The latter includes every token that is not classified as any of the previous three classes. As this covers a wide variety of tokens (including named entities that do not belong to one of the three aforementioned classes, non-named entities, verbs, stopwords, etc.), we ignore the OTHR label during the evaluation. That is, we only take into account the classification scores of LOC, ORG, and PER classes.

\begin{table}[htb]
    \centering
	\caption{The collected datasets for NER task and their properties: the number of sentences and tagged words.}
	\label{tab:ner-datasets}
	\begin{tabular}{llrrcl}
		Language	& Dataset & Sentences & Tags \\ 
		\hline
		Croatian	& hr500k & $24794$ & $28902$ \\ 
		English 	& CoNLL-2003 NER & $20744$ & $43979$ \\ 
		Estonian	& Estonian NER corpus & $14287$ & $20965$ \\ 
		Finnish  	& FiNER data & $14484$ & $16833$ \\ 
		Latvian  	& LV Tagger train data & $9903$ & $11599$ \\ 
 		Lithuanian	& TildeNER & $5500$ & $7000$ \\ 
		Russian 	& factRuEval-2016 & $4907$ & $9666$ \\ 
		Slovene\tablefootnote{The Slovene ssj500k originally contains more sentences, but only $9489$ are annotated with named entities.} 	& ssj500k & $9489$ & $9440$ \\ 
		Swedish	& Swedish NER & $9369$ & $7292$ \\ 
		\hline
	\end{tabular}
\end{table}

\subsubsection{POS-tagging and Dependency parsing}
We used datasets in nine languages (Croatian, English, Estonian, Finnish, Latvian, Lithuanian, Russian, Slovene and Swedish) to test models on the POS-tagging and DP tasks. The datasets are obtained from the Universal Dependencies 2.3 \citep{universal-dependencies}, except the Lithuanian ALKSNIS dataset, which comes from the Universal Dependencies 2.8. The number of sentences and tokens is shown in Table~\ref{tab:depparse-datasets}. We used 17 Universal POS tags for the POS-tagging task as they are the same in all languages and did not predict language-specific XPOS tags.

\begin{table}[htb]
    \centering
	\caption{POS-tagging and dependency parsing datasets and their properties: the treebank, number of sentences, number of tokens, and information about the size of the splits. 
	}
	\label{tab:depparse-datasets}
	\begin{tabular}{llrrrrr}
	Language & Treebank & Tokens & Sentences & Train & Validation & Test \\
	\hline
	Croatian & SET & $197044$ & $8889$ & $6983$ & $849$ & $1057$ \\
	English & EWT & $254854$ & $16622$ & $12543$ & $2002$ & $2077$ \\
	Estonian & EDT & $434245$ & $30723$ & $24384$ & $3125$ & $3214$ \\
	Finnish & TDT & $202208$ & $15136$ & $12217$ & $1364$ & $1555$ \\
	Latvian & LVTB & $152706$ & $9920$ & $7163$ & $1304$ & $1453$ \\
	Lithuanian & ALKSNIS & 70051 & 3642 & 2341 & 617 & 684 \\
	Russian & GSD & $99389$ & $5030$ & $3850$ & $579$ & $601$ \\
	Slovene & SSJ & $140670$ & $8000$ & $6478$ & $734$ & $788$ \\
	Swedish & Talbanken & $96858$ & $6026$ & $4303$ & $504$ & $1219$ \\
	\hline
	\end{tabular}
\end{table}

We use two evaluation metrics in the dependency parsing task, the mean of unlabeled and labelled attachment scores (UAS and LAS) on the test set. The UAS and LAS are standard accuracy metrics in dependency parsing. The UAS score is defined as the proportion of tokens that are assigned the correct syntactic head. The LAS score is the proportion of tokens that are assigned the correct syntactic head and the correct dependency label \citep{speech-and-language-processing}.

\subsubsection{CoSimLex}
\label{sec:cosimlexIntro}
In contrast to other datasets which are used to evaluate the performance of embeddings on specific tasks, the CoSimLex task \citep{ArmendarizEtAl20LREC} allows direct investigation of embeddings' properties. CoSimLex contains pairs of words and their similarity ratings assigned by human annotators. The crucial difference to previous such datasets is that the words appeared within a short text (context) when presented to the human annotators. Therefore, the word similarity ratings take the context into account, making the dataset suitable to evaluate the contextualised embeddings. The dataset is based on pairs of words from SimLex-999 \citep{HillEtAl15SimLex} to allow comparison with the context-independent case. 
CoSimLex consists of 340 word-pairs in English, 112 in Croatian, 111 in Slovene, and 24 in Finnish. Each pair is rated within two different contexts, giving a total of 1174 scores of contextual similarity. 

As the example in \Cref{fig:cosimlex_example} shows, for each pair of words, two different contexts are presented in which these two words appear. The words in contexts produce two similarity scores, each related to one of the contexts, calculated as the mean of all the annotators' ratings for that context. This is accompanied by two standard deviation scores. 
and the four inflected forms of the words exactly as they appear in the contexts. 
Note that in the morphologically rich languages (such as Slovene, Croatian, and Finnish), many inflections of the two words are possible. 

\begin{figure}[!ht]
\centering
\caption{An example from the English CoSimLex, showing a word pair with two contexts, each with the mean and standard deviation of human similarity judgements. The original SimLex values for the same word pair without context are shown for comparison. The p-Value shown results from the Mann-Whitney U test for similarity of distributions, showing that the human judgements differ significantly between the two contexts.}
\label{fig:cosimlex_example}

\resizebox{\columnwidth}{!}{%
\begin{tabular}{|llr|} \hline 
\bf Word1: man & 
\bf Word2: warrior &
\textbf{SimLex}: $\mu$ 4.72 $\sigma$ 1.03 \\ \hline

\bf Context1 & 
 & 
\textbf{Context1:} $\mu$ 7.88 $\sigma$ 2.07 \\ 
\multicolumn{3}{|p{\linewidth}|}{
When Jaimal died in the war, Patta Sisodia took the command, but he too died in the battle. These young \textbf{men} displayed true Rajput chivalry. Akbar was so impressed with the bravery of these two \textbf{warriors} that he commissioned a statue of Jaimal and Patta riding on elephants at the gates of the Agra fort.
}\\ \hline
\bf Context2 & 
 & 
\textbf{Context2:} $\mu$ 3.27 $\sigma$ 2.87 \\
\multicolumn{3}{|p{\linewidth}|}{
She has a dark past when her whole family was massacred, leaving her an orphan. By day, Shi Yeon is an employee at a natural history museum. By night, she's a top-ranking woman \textbf{warrior} in the Nine-Tailed Fox clan, charged with preserving the delicate balance between \textbf{man} and fox.
}\\ \hline
\bf  &
\bf  &
\textbf{p-Value:} $1.3\times10^{-6}$ \\ \hline
\end{tabular}
}%

\end{figure}

Model performance is evaluated using two metrics, which measure different aspects of prediction quality:
\begin{description}
\item [M1 - Predicting Changes:] The first metric measures the ability of a model to predict the \emph{change in similarity ratings between the two contexts} for each word pair. This is evaluated via the correlation between the changes predicted by the system and those derived from human ratings, using the uncentered Pearson correlation. This gives a measure of the accuracy of predicting the relative magnitude of changes and allows for differences in scaling while maintaining the effect of the direction of change. The standard centered correlation normalises on the mean, so it could give high values even when a system predicts changes in the wrong direction, but with a similar distribution over examples.

\[ M1 = CC_{uncentered}= \frac{\sum_{i=1}^{n} (x_i)(y_i)}{\sqrt{(\sum_{i=1}^{n} x_i)^2(\sum_{i=1}^{n} y_i)^2}} \]

\item[M2 - Predicting Ratings:] The second metric measures the ability to predict the absolute similarity rating for each word pair in each context. This was evaluated using the harmonic mean of the Pearson and the Spearman correlation with gold-standard human judgements.
\end{description}

\subsubsection{Monolingual and cross-lingual analogies}
The word analogy task ($x$ is to $y$ as $a$ is to $b$) was popularised by \citet{mikolov2013distributed}. The goal is to find a term $y$ for a given term $x$ so that the relationship between $x$ and $y$ best resembles the given relationship $a : b$. In the used datasets, There are two main groups of categories: semantic and syntactic. To illustrate a semantic relationship (country and its capital), consider, for example, that the word pair $a : b$ is given as ``Finland : Helsinki''. The task is to find the term $y$ corresponding to the relationship ``Sweden : $y$'', with the expected answer being $y=$ Stockholm. In syntactic categories, each category refers to a grammatical feature, e.g., adjective degrees of comparison. The two words in any given pair then have a common stem (or even the same lemma); e.g., given the word pair ``long : longer'', we have an adjective in its base form and the same adjective in the comparative form. The task is to find the term $y$ corresponding to the relationship ``dark : $y$'', with the expected answer being $y=$ darker, i.e. a comparative form of the adjective dark. 

In the vector space, the analogy task is transformed into vector arithmetic. We search for nearest neighbours, i.e. we compute the distance $d$ between vectors: $d$(vec(Finland), vec(Helsinki)) and search for word $y$ which would give the closest result in distance $d$(vec(Sweden), vec($y$)). 
We use the monolingual and cross-lingual analogy datasets in nine languages (Croatian, English, Estonian, Finnish, Latvian, Lithuanian, Russian, Slovenian, and Swedish) \citep{ulcar-2020-multilingual}. Here, the analogies are already prespecified so we do not search for the closest result but only check if the prespecified word is indeed the closest; alternatively, we measure the distance between the given pairs. 
The proportion of correctly identified words in the five nearest vectors forms a statistic called accuracy@5, which we report as the result.

In the cross-lingual setting, for two languages $L_1$ and $L_2$, the word analogy task  matches each relation in one language with each relation from the same category in the other language. For cross-lingual contextual mappings, the presented word analogy task is not well-suited as it only contains words without their context. We describe our approach for applying this task in \Cref{sec:evalSettingsAnalogies}.

\subsubsection{SuperGLUE tasks}
\label{SuperGLUEbenchmark}
SuperGLUE (Super General Language Understanding Evaluation) \citep{wang2019superglue} is a benchmark for testing natural language understanding (NLU) of models. It is styled after the GLUE benchmark \citep{wang2019glue}, but much more challenging. It provides a single-number metric for each of its tasks that enables the comparison and progress of NLP models. The tasks are diverse and comprised of question answering (BoolQ, COPA, MultiRC, and ReCoRD tasks), natural language inference (CB and RTE tasks), coreference resolution (WSC), and word sense disambiguation (WiC). Non-expert humans evaluated all the tasks to give a human baseline to machine systems. 
Please refer to the original paper for an extensive description of the tasks.

To evaluate cross-lingual transfer and test specifics of morphologically rich languages, we translated the SuperGLUE datasets to Slovene. We  used partially human translation (HT) and partially machine translation (MT). The details are presented in Table \ref{tab:superglue_translation}. Some datasets are too large (BoolQ, MultiRC, ReCoRD, RTE) to be fully human translated with our budget. We thus provide ratios between the human translated and the original English sizes. For MT from English to Slovene, we used the GoogleMT Cloud service. In our evaluation, we use six of the original eight tasks. 

The WSC dataset cannot be machine-translated because it requires human assistance and verification. First, GoogleMT translations cannot handle the correct placement of HTML tags indicating coreferences. 
The second reason is that in Slovene coreferences can also be expressed with verbs, while coreferences in English are mainly nouns, proper names and pronouns. 
This makes the task more difficult in Slovene compared to English because solutions cover more  types of words. 

We did not include ReCoRD in the Slovene benchmark due to the low quality of the resulting dataset, consisting of confusing and ambiguous examples. Besides imperfect translations, there are differences between English and Slovene ReCoRD tasks due to the morphological richness of Slovene. In Slovene, the correct declension of a query is often not present in the text, making it impossible to provide the correct answer. Finally, similarly to WSC, ReCoRD is also affected by the problem of translating HTML tags with GoogleMT. 

The WiC task cannot be translated and would have to be conceived anew because it is impossible to transfer the same set of meanings of a given word from English to a target language. 

\begin{table}[htb]
\caption{The number of instances in the original English and translated Slovene SuperGLUE tasks. HT stands for human translation and MT for machine translation. The ``ratio`` indicates the ratio between the number of human translated instances and all instances. }
\label{tab:superglue_translation}
\begin{center}
\scalebox{0.9}{
\begin{tabular}{l|lllll}
\textbf{Dataset} & \textbf{split} & \textbf{English} & \textbf{HT} & \textbf{ratio} & \textbf{MT} \\ \hline
\textbf{BoolQ} & train & 9427 & 92 & 0.0098 & yes \\
\textbf{} & val & 3270 & 18 & 0.0055 & yes \\
\textbf{} & test & 3245 & 30 & 0.0092 & yes \\ \hline
\textbf{CB} & train & 250 & 250 & 1.0000 & yes \\
\textbf{} & val & 56 & 56 & 1.0000 & yes \\
\textbf{} & test & 250 & 250 & 1.0000 & yes \\ \hline
\textbf{COPA} & train & 400 & 400 & 1.0000 & yes \\
\textbf{} & val & 100 & 100 & 1.0000 & yes \\
\textbf{} & test & 500 & 500 & 1.0000 & yes \\ \hline
\textbf{MultiRC} & train & 5100 & 15 & 0.0029 & yes \\
\textbf{} & val & 953 & 3 & 0.0031 & yes \\
\textbf{} & test & 1800 & 30 & 0.0167 & yes \\ \hline
\textbf{ReCoRD} & train & 101000 & 60 & 0.0006 & / \\
\textbf{} & val & 10000 & 6 & 0.0006 & / \\
\textbf{} & test & 10000 & 30 & 0.0030 & / \\ \hline
\textbf{RTE} & train & 2500 & 232 & 0.0928 & yes \\
\textbf{} & val & 278 & 29 & 0.1043 & yes \\
\textbf{} & test & 300 & 29 & 0.0967 & yes \\ \hline
\textbf{WiC} & train & 6000 & / & / & / \\
\textbf{} & val & 638 & / & / & / \\
\textbf{} & test & 1400 & / & / & / \\ \hline
\textbf{WSC} & train & 554 & 554 & 1.0000 & / \\
\textbf{} & val & 104 & 104 & 1.0000 & / \\
\textbf{} & test & 146 & 146 & 1.0000 & /
\end{tabular}
}
\end{center}
\end{table}

\subsubsection{Terminology alignment}
\label{terminologyBenchmark}
Terms are single words or multi-word expressions denoting concepts from specific subject fields. The bilingual terminology alignment task aligns terms between two candidate term lists in two different languages. The primary purpose of terminology alignment is to build a bilingual term bank - i.e. a list of terms in one language and their equivalents in another language. 

Given a pair of terms "$t_1$" and "$t_2$", where $t_1$ is from one language and $t_2$ is its equivalent from the second language, we measured the cosine distance between vector of $t_1$ and vectors of all terms from the second language. If the vector of $t_2$ is the closest to $t_1$ among all terms, we count the pair as correctly aligned. For example, for a pair of terms from the Slovenian-English term bank ``računovodstvo - accounting'', we map the Slovene word embedding of the word ``računovodstvo'' from Slovene to English and check among all English word vectors for the vector that is the closest to the mapped Slovenian vector for ``računovodstvo''. If the closest vector is ``accounting'', we count this as a success, else we do not. This measure is called accuracy@1 score or 1NN score: the number of successes, divided by the number of all examples, in this case, dictionary pairs. A similar measure accuracy@n checks the proportion of correct translations among $n$ closest words.
In this work, we use the term alignment task to compare different embedding models. 

For building contextualised vector representations of terms, we used the Europarl corpus \citep{koehn2005europarl,tiedmann2012}. For Croatian, Europarl is not available, so we used DGT translation memory \citep{steinberger2013dgt} instead. We used these corpora, composed of mostly EU legislation texts available in all EU languages, to create contextual word embeddings. For single-word terms, we represent each term as the average vector of all contextual vector representations for that word, found in the corpus. For multi-word terms, we used a two-step approach. If the term appears in the corpus, we represent each term occurrence as the average vector of the words it is composed of. We then average over all the occurrences, as with single-word terms. In case the term does not appear in the corpus, we represent it as the average of all words it is composed of, where word vectors are averaged over all occurrences in the corpus.

To evaluate the performance of embedding-based terminology alignment, we used Eurovoc \citep{steinberger2002cross}, a multilingual thesaurus with more than 10,000 terms available in all EU languages. The models were evaluated for the following pairs of languages: Croatian-Slovenian, Estonian-Finnish, Latvian-Lithuanian, and English paired with each of the following: Croatian, Estonian, Finnish, Latvian, Lithuanian, Slovenian, and Swedish. For each language pair, we evaluated the terminology alignment in both directions, i.e. we took terms from the first language and searched for the closest terms in the second language, then repeated the procedure by taking the second language terms and searched for the closest terms in the first language.

\subsection{Evaluation settings}
\label{sec:evalSettings}
We split our evaluations into two categories: monolingual and cross-lingual. In the monolingual evaluation, we compare fastText, ELMo,  monolingual BERT-like models (English, Russian, Finnish, Swedish, Slovene, Croatian, Estonian, and Latvian), trilingual BERT models (FinEst, CroSloEngual, and LitLat BERT), and massively multilingual BERT models (mBERT and XLM-R). The exact choice of compared models depends on the availability of datasets in specific languages. In the cross-lingual setting, we compare cross-lingual maps for ELMo models, massively multilingual BERT models, and trilingual BERT models. The specifics of models for individual tasks are described below.

\subsubsection{Named entity recognition}
For each of the compared embeddings, we tested a separate neural architecture, adapted to the specifics of the embeddings. For fastText and ELMo embeddings, we trained NER classifiers by inputting word vectors for each token in a given sentence, along with their labels. We used a model with two bidirectional LSTM layers with 2048 units. On the output, we used the time-distributed softmax layer. For ELMo embeddings, we computed a weighted average of the three embedding vectors for each token, by learning the weights during the training. We used the Adam optimizer with a learning rate of $10^{-4}$ and trained for 5 epochs.

For BERT models, we fine-tuned each model on the NER dataset for 3 epochs. We used the code by HuggingFace\footnote{\href{https://github.com/huggingface/transformers/tree/master/examples/legacy/token-classification}{https://github.com/huggingface/transformers/tree/master/examples/legacy/token-classification}} for NER classification.

\subsubsection{POS-tagging}
For training POS-tagging classifiers with fastText or ELMo embeddings, we used the same approach and hyper-parameters as described above for NER, but a different neural network architecture. We trained models with four hidden layers, three bidirectional LSTMs and one fully connected feed-forward layer. The three LSTM layers have 512, 512, and 256 units, respectively. The fully connected layer has 64 neurons. 

For BERT models, we fine-tuned each model for 3 epochs, using the POS classification code by HuggingFace as for NER.

\subsubsection{Dependency parsing}
To train dependency parsers using ELMo embeddings, we used the SuPar tool by Yu Zhang\footnote{\href{https://github.com/yzhangcs/parser}{https://github.com/yzhangcs/parser}}. SuPar is based on the deep biaffine attention \citep{dozat2017deepbiaffine}. We modified the SuPar tool to accept ELMo embeddings on the input; specifically, we used the concatenation of the three ELMo vectors. The modified code has been made publicly available\footnote{\href{https://github.com/EMBEDDIA/supar-elmo}{https://github.com/EMBEDDIA/supar-elmo}}. We trained the parser for 10 epochs for each language, using separately L-ELMo and EFML embeddings.

For fine-tuning BERT models, we modified the dep2label-bert tool \citep{strzyz-etal-2019-viable,gomez-rodriguez-etal-2020-unifying} to work with newer versions of HuggingFace's transformers library and to support both RoBERTa and BERT-based models. We used the modified tool to fine-tune all the BERT/RoBERTa models on the dependency parsing task for 10 epochs. We used the arc-Standard algorithm in transition-based sequence labelling encoding. The modified tool is publicly available\footnote{\href{https://github.com/EMBEDDIA/dep2label-transformers}{https://github.com/EMBEDDIA/dep2label-transformers}}.


\subsubsection{Analogies}
\label{sec:evalSettingsAnalogies}
The word analogy task was initially designed for static embeddings. To evaluate contextual embeddings, we have to use the words of each analogy entry in a context. Such contexts may not exist in general corpora for some categories. We used a boilerplate sentence "If the term [w1] corresponds to the term [w2], then the term [w3] corresponds to the term [w4]." Here, [w1] through [w4] represent the four words from an analogy entry. We translated the boilerplate sentence to every language where a suitable analogy dataset is available (Croatian, English, Estonian, Finnish, Latvian, Lithuanian, Russian, Slovene, Swedish) \citep{ulcar-2020-multilingual}. 

For ELMo models, we concentrated on evaluating cross-lingual mapping approaches. Given a cross-lingual analogy entry (i.e. the first two words in one language, and the last two words in another language), we filled the boilerplate sentence in the training language with the four analogy words (two of them being in the "wrong" language) and extracted the vectors for words "w1" and "w2". We then filled the boilerplate sentence in the testing language with the same four words and extracted the vectors for words "w3" and "w4". We evaluated the quality of the mapping by measuring the distance between vector $v(w_4)$ and vector $v(w_2)-v(w_1)+v(w_3)$.

BERT models are masked language models, so we tried to exploit that in this task. We masked the word "w2" and tried to predict it, given every other word. In the cross-lingual setting, the sentence after the comma and the words "w3" and "w4" were therefore given in the source/training language, while the sentence before the comma and word "w1" were given in target/evaluation language. The prediction for the masked word "w2" was expected in the target/testing language, as well.

\subsubsection{SuperGLUE}
We fine-tuned BERT models on SuperGLUE tasks using the Jiant tool \citep{phang2020jiant}. We used a single-task learning setting for each task and fine-tuned them for 100 epochs each, with the initial learning rate of $10^{-5}$. Each model was fine-tuned using either machine translated or human translated datasets of the same size. 

\subsubsection{Terminology alignment}

For ELMo embeddings, we concatenated the three ELMo vectors into one 3072-dimensional vector for each term. For BERT models, we extracted the vectors from outputs of the last 4 layers and concatenated them to produce a 3072-dimensional vector for each term.  

\section{Results}
\label{sec:results}
We present two sets of results. First, in \Cref{monoEvaluation} we evaluate monolingual models, followed by evaluation of cross-lingual transfer in \Cref{crossEvaluation}.

\subsection{Monolingual evaluations}
\label{monoEvaluation}
The monolingual evaluation is split into six subsections according to the type of task. We start with NER, followed by POS-tagging, dependency parsing, CoSimLex, analogies, and SuperGLUE tasks. The results shown for classification tasks NER, POS-tagging, and dependency parsing are the averages of five individual evaluation runs.

\subsubsection{Named entity recognition}
In \Cref{tab:results-elmo-monolingual-ner}, we present the results of fastText non-contextual baseline, compared with two types of contextual ELMo embeddings, ELMoForManyLangs and L-ELMo (described in \Cref{sec:ELMo}). L-ELMo, trained on much larger datasets, is the best in every language except Latvian. 
The fastText baseline lags behind both ELMo embeddings. 

\begin{table}[!h]
\begin{center}
\caption{The comparison of fastText non-contextual baseline with two types of ELMo embeddings, EFML and L-ELMo on the NER task.  The results are given as macro $F_1$ scores. The best model for each language is in \textbf{bold}. There is no Lithuanian EFML model.}
\label{tab:results-elmo-monolingual-ner}
\begin{tabular}{lcc|c}
      Language & fastText & EFML & L-ELMo \\
      \hline
        Croatian   & 0.570 & 0.733 & \textbf{0.810} \\
        English    & 0.807 & 0.879 & \textbf{0.922} \\
        Estonian   & 0.734 & 0.828 & \textbf{0.895} \\
        Finnish    & 0.692 & 0.882 & \textbf{0.923} \\
        Latvian    & 0.557 & \textbf{0.838} & 0.818 \\
        Lithuanian & 0.359 & N/A & \textbf{0.755} \\ 
        Slovenian  & 0.478 & 0.772 & \textbf{0.849} \\
        Swedish    & 0.663 & 0.829 & \textbf{0.852} \\
      \hline
\end{tabular}
 \end{center}
\end{table}

The results of BERT models are presented in \Cref{tab:results-bert-monolingual-ner}. Each of the listed BERT models was fine-tuned on NER datasets in languages where that makes sense: monolingual and trilingual BERT models were used in languages used in their pretraining, and massively multilingual models (mBERT and XLM-R) were fine-tuned for all used languages.  

\begin{table}[!htb]
\begin{center}
 \caption{The results of NER evaluation task for multilingual BERT (mBERT), XLM-RoBERTa (XLM-R), trilingual BERT-based (TRI) and monolingual BERT-based (MONO) models. The scores are macro average $F_1$ scores of the three named entity classes.}
\label{tab:results-bert-monolingual-ner}
\begin{tabular}{lrrrr}
      Language & mBERT & XLM-R & TRI & MONO\\
      \hline
      Croatian & 0.801 & 0.833 & \textbf{0.886} & 0.881 \\ 
      English & 0.938 & 0.941 & \textbf{0.944} & 0.943 \\ 
      Estonian & 0.900 & 0.913 & 0.930 & \textbf{0.936} \\ 
      Finnish & 0.934 & 0.932 & \textbf{0.957} & 0.952 \\ 
      Latvian & 0.847 & 0.859 & \textbf{0.863} & 0.145 \\ 
      Lithuanian & 0.833 & 0.802 & \textbf{0.863} & - \\ 
      Slovenian & 0.885 & 0.912 & 0.928 & \textbf{0.933} \\ 
      Swedish & 0.844 & 0.875 & - & \textbf{0.887} \\ 
      \hline
\end{tabular}
\end{center}
\end{table}

All BERT-like models perform similarly in English. XLM-R outperforms mBERT on all languages, except Lithuanian and Finnish. Trilingual models outperform both mBERT and XLM-R on all languages and outperform most monolingual models, except Est-RoBERTa on Estonian and SloBERTa on Slovenian. The monolingual LVBERT model performs poorly, which is an indication that this model was not trained on a large enough dataset. The best performing trilingual model in English is CroSloEngual BERT. On Estonian, Est-RoBERTa ($F_1=0.936$) significantly outperforms EstBERT ($F_1=0.870$). Again, the latter does not seem to be trained on a large enough dataset.  
Comparing BERT results in \Cref{tab:results-bert-monolingual-ner} with ELMo results in \Cref{tab:results-elmo-monolingual-ner}, we can observe clear dominance of BERT models. The extracted ELMo embedding vectors are clearly not competitive to the entire pretrained BERT models on the NER task.

\subsubsection{POS-tagging}
In \Cref{tab:results-elmo-monolingual-pos}, we present the results of fastText non-contextual baseline, compared with two types of contextual ELMo embeddings, EFML and L-ELMo. Again, L-ELMo models are the best in all languages. Some results are surprisingly low, but this is the effect of the low quality of the datasets.

\begin{table}[!ht]
\begin{center}
\caption{The comparison of fastText non-contextual baseline with two types of ELMo embeddings (EFML and L-ELMo) on the POS-tagging task. The results are given as micro $F_1$ scores. The best results for each language are in \textbf{bold}. There is no Lithuanian EFML model.}
\label{tab:results-elmo-monolingual-pos}
\begin{tabular}{lcc|c}
      Language & fastText & EFML & L-ELMo \\
      \hline
        Croatian   & 0.512 & 0.573 & \textbf{0.963} \\
        English    & 0.769 & 0.603 & \textbf{0.952} \\
        Estonian   & 0.640 & 0.508 & \textbf{0.969} \\
        Finnish    & 0.506 & 0.389 & \textbf{0.966}  \\
        Latvian    & 0.462 & 0.489 & \textbf{0.940}  \\
        Lithuanian & 0.303 & N/A & \textbf{0.316} \\
        Russian    & 0.518 & 0.349 & \textbf{0.929} \\
        Slovenian  & 0.527 & 0.541 & \textbf{0.966} \\
        Swedish    & 0.275 & 0.313 & \textbf{0.933} \\
      \hline
\end{tabular}
 \end{center}
\end{table}

\begin{table}[!ht]
\begin{center}
 \caption{The results of POS-tagging evaluation task for multilingual BERT (mBERT), XLM-RoBERTa (XLM-R), trilingual BERT-based (TRI) and monolingual BERT-based (MONO) models expressed with $F_1$ scores. The best results for each language are in \textbf{bold}.
 }
\label{tab:results-bert-monolingual-pos}
\begin{tabular}{lrrrr}
      Language & mBERT & XLM-R & TRI & MONO\\
      \hline
      Croatian & 0.978 & 0.981 & \textbf{0.982} & 0.981 \\ 
      English & 0.964 & \textbf{0.972} & 0.968 & 0.967 \\  
      Estonian & 0.966 & 0.970 & 0.973 & \textbf{0.977} \\
      Finnish & 0.961 & 0.977 & 0.976 & \textbf{0.980} \\ 
      Latvian & 0.946 & 0.960 & \textbf{0.966} & 0.048 \\ 
      Lithuanian & 0.934 & \textbf{0.964} & 0.961 & - \\
      Russian & 0.974 & \textbf{0.976} & - & 0.975 \\
      Slovenian & 0.984 & 0.988 & 0.990 & \textbf{0.991} \\
      Swedish & 0.979 & 0.981 & - & \textbf{0.988} \\
      \hline
\end{tabular}
\end{center}
\end{table}

The results of BERT models are presented in \Cref{tab:results-bert-monolingual-pos}. Each of the listed BERT models was fine-tuned on POS datasets in languages where that makes sense: monolingual and trilingual BERT models were used in languages used in their pretraining, and massively multilingual models (mBERT and XLM-R) were fine-tuned for all used languages.

The results show that trilingual models and massively multilingual BERT models are very competitive in the POS-tagging task, differences being relatively small and language-dependent. Nevertheless, for some languages the same pattern appears as in NER: in Slovenian and Estonian monolingual models are the best again, the trilingual models outperform monolingual Croatian and English models, the monolingual Latvian BERT model, which was pretrained on insufficient amounts of data, performs poorly again. The Est-RoBERTa model outperforms EstBERT again (0.977 vs. 0.961), all trilingual models score the same on English. Comparing BERT and ELMo results in Tables \ref{tab:results-elmo-monolingual-pos} and \Cref{tab:results-bert-monolingual-pos}, we again observe a clear dominance of BERT models.

\subsubsection{Dependency parsing}
In \Cref{tab:results-elmo-monolingual-dp}, we compare two types of contextual ELMo embeddings (EFML and L-ELMo) on the dependency parsing task. L-ELMo models outperform EFML on all languages. The difference between them is very small in English and Russian, while the largest difference occurs in Slovenian and Estonian. 

\begin{table}[!ht]
\begin{center}
\caption{The comparison of two types of ELMo embeddings (EFML and L-ELMo) on the dependency parsing task. Results are given as UAS and LAS scores. The best results for each language are typeset in \textbf{bold}. There is no Lithuanian EFML model.}
\label{tab:results-elmo-monolingual-dp}
\begin{tabular}{lcc|cc}
       & \multicolumn{2}{c|}{ELMoForManyLangs} & \multicolumn{2}{c}{L-ELMo} \\
      Language & UAS & LAS & UAS & LAS \\
      \hline
        Croatian   & 88.18 & 79.45 & \textbf{91.74} & \textbf{85.84} \\
        English    & 90.28 & 86.29 & \textbf{90.53} & \textbf{87.16} \\
        Estonian   & 81.19 & 72.50 & \textbf{89.54} & \textbf{85.45}  \\
        Finnish    & 88.27 & 83.44 & \textbf{90.83} & \textbf{86.86}  \\
        Latvian    & 87.17 & 80.76 & \textbf{88.85} & \textbf{82.82}  \\
        Lithuanian & - & - & \textbf{82.84} & \textbf{72.16} \\
        Russian    & 89.28 & 83.29 & \textbf{89.33} & \textbf{83.54} \\
        Slovenian  & 85.55 & 77.73 & \textbf{93.70} & \textbf{91.39} \\
        Swedish    & 88.03 & 83.09 & \textbf{89.70} & \textbf{85.07} \\
      \hline
\end{tabular}
 \end{center}
\end{table}

\begin{table}[!ht]
\begin{center}
 \caption{The results of the evaluation in the dependency parsing task for multilingual BERT (mBERT), XLM-RoBERTa (XLM-R), trilingual BERT-based (TRI) and monolingual BERT-based (MONO) models.  The results are given as LAS scores. The best results for each language are typeset in \textbf{bold}.
  }
\label{tab:results-bert-monolingual-dp1}
\begin{tabular}{lrrrr}
      Language & mBERT & XLM-R & TRI & MONO \\
      \hline
      Croatian & 70.38 & 78.39 & \textbf{82.36} & - \\ 
      English & 83.19 & \textbf{84.94} & 83.91 & 83.55 \\ 
      Estonian & 56.27 & 68.91 & 75.67 & \textbf{78.64} \\ 
      Finnish & 57.22 & 71.12 & 79.96 & \textbf{83.64} \\
      Latvian & 54.61 & 69.26 & \textbf{74.32} & 56.87 \\
      Lithuanian & 44.08 & 56.61 & \textbf{61.66} & - \\
      Russian & 70.00 & 73.47 & - & \textbf{80.90} \\
      Slovenian & 68.08 & 79.27 & \textbf{85.38} & 84.41 \\
      Swedish & 74.04 & 80.93 & - & \textbf{85.83} \\
      \hline
\end{tabular}
\end{center}
\end{table}

The results of BERT models are presented in \Cref{tab:results-bert-monolingual-dp1}. Each of the listed BERT models was fine-tuned on POS datasets in languages where that makes sense: monolingual and trilingual BERT models were used in languages used in their pretraining, and massively multilingual models (mBERT and XLM-R) were fine-tuned for all used languages.

The results show that the differences between monolingual, trilingual, and massively multilingual BERT models are language-dependent. CroSloEngual is again the best performing trilingual model in English. EstBERT performs well on this task, but still worse than Est-RoBERTa (77.44 and 78.64 LAS score, respectively). 

Surprisingly, comparing BERT and ELMo results in Tables \ref{tab:results-elmo-monolingual-dp} and \Cref{tab:results-bert-monolingual-dp1}, shows that L-ELMo models dominate in all languages. These results indicate that BERT models shall not always be the blind choice in text classification, as ELMo might still be competitive in some tasks. 

\subsubsection{CoSimLex}
In \Cref{tab:results-metric1-monolingual-CoSimLex}, we compare performance on the CoSimLex word similarity in context task for two types of ELMo models (EFML and L-ELMo) and BERT models (massively multilingual mBERT and two trilingual models: CroSloEngual BERT and FinEst BERT). The performance is expressed with two metrics: M1 measures the ability to predict the change in similarity due to a change in context, measured as uncentered Spearman correlation between the predicted and actual change of similarity scores; and M2 measures the ability to predict absolute ratings of similarity in context, measured as the harmonic mean of the Spearman and Pearson correlations between predicted and actual similarity scores.  See \Cref{sec:cosimlexIntro} for details.


\begin{table}[!ht]
\caption{Comparison of different ELMo and BERT embeddings on the CoSimLex datasets. We compare performance via the uncentered Spearman correlation between the predicted and true change in similarity scores (M1), and the harmonic mean of the Spearman and Pearson correlations between predicted and true similarity scores (M2). Trilingual and monolingual models are based on either \textsuperscript{a} the original BERT model, \textsuperscript{b} the Electra approach or \textsuperscript{c} the RoBERTa model.} 
\label{tab:results-metric1-monolingual-CoSimLex}
\begin{center}
\resizebox{\linewidth}{!}{
\begin{tabular}{ll|ll|llll}
\multicolumn{2}{c}{ }& \multicolumn{2}{c}{ELMo models} & \multicolumn{4}{c}{BERT Models} \\
Model    & Metric & EFML & L-ELMo & mBERT & XLM-R & TRI & MONO  \\ \hline
English  & M1 & 0.556 & \bf 0.570 & 0.713 & 0.545 & 0.719\textsuperscript{a} & \bf 0.729\textsuperscript{a} \\
Croatian & M1 & 0.520 & \bf 0.662 & 0.587 & 0.444 & \bf 0.715\textsuperscript{a} & 0.351\textsuperscript{b} \\
Slovene  & M1 & 0.467 & \bf 0.550 & 0.603 & 0.440 & \bf 0.673\textsuperscript{a} & 0.574\textsuperscript{c} \\
Finnish  & M1 & 0.403 & \bf 0.452 & 0.671 & 0.260 & \bf 0.672\textsuperscript{a} & 0.595\textsuperscript{a} \\ \hline
English  & M2 & 0.449 & \bf 0.510 & 0.573 & 0.440 & 0.601\textsuperscript{a} & \bf 0.653\textsuperscript{a} \\
Croatian & M2 & 0.433 & \bf 0.529 & 0.443 & 0.387 & \bf 0.642\textsuperscript{a} & 0.391\textsuperscript{b} \\
Slovene  & M2 & 0.328 & \bf 0.516 & 0.516 & 0.355 & \bf 0.589\textsuperscript{a} & 0.445\textsuperscript{c} \\
Finnish  & M2 & 0.403 & \bf 0.407 & 0.289 & 0.053 &  0.533\textsuperscript{a} & \bf 0.570\textsuperscript{a} \\ \hline
\end{tabular}}
\end{center}
\end{table}

Among ELMo models, L-ELMo models consistently outperform EFML models, producing closer scores to humans in both metrics and all four languages.  Among BERT models, the trilingual models (see \Cref{sec:ourBERT}) do best for most languages and metrics. The exceptions are English, in which the original monolingual BERT outperforms the trilingual models in both metrics, and the monolingual Finnish model (FinBERT), which achieves the best results for M2. We note that these best-performing models significantly outperform the standard multilingual BERT (mBERT) in all cases except M1 for Finnish, where mBERT's results are similar.

Comparing ELMo and BERT models, BERT models are more successful and predict similarities closer to human assigned scores. Interestingly, looking at the different types of BERT models, the ones that are based on the original BERT model seem to do much better than more recent variants. Multilingual mBERT does significantly better than XLM-RoBERTa, and the best model for every category is based on the original BERT formulation (English BERT, CroSloEngual, FinEst and FinBERT). The monolingual Slovene model (SloBERTa) performs poorly (it is based on the same RoBERTa variant as XLM-RoBERTa). The monolingual Croatian model (BERTić), trained using the Electra approach (see \Cref{sec:monoBERTs}), does especially poor on this task. It seems possible that the Electra and RoBERTa approaches, due to their different pretraining objectives, produce less human-like models in terms of their embedding similarities - but further experiments are required to draw stronger conclusions. 

\subsubsection{Analogies}

In \Cref{tab:results-elmo-monolingual-analogy}, we present the comparison between two types of ELMo embeddings (EFML and L-ELMo) on the word analogy task. We used two distance metrics to measure the distance between the expected and the actual result. The results strongly depend on the used distance metric. Using the Euclidean distance, EFML outperforms L-ELMo on five languages, and L-ELMo wins in three languages. On these three languages (Croatian, Estonian, and Slovenian), results of EFML are significantly worse than on other languages. Using cosine distance, which is more suited for high-dimensional vector spaces, L-ELMo outperforms EFML on all languages.

\begin{table}[!ht]
\begin{center}
\caption{The comparison of two types of ELMo embeddings (EFML and L-ELMo) on the word analogy task. Results are reported as the macro average distance between expected and actual word vector of the word w4. Two distance metrics were used: cosine (Cos) and Euclidean (Euc). The best results (shortest distance) for each language and metric are typeset in \textbf{bold}. There is no Lithuanian EFML model.}
\label{tab:results-elmo-monolingual-analogy}
\begin{tabular}{lcc|cc}
       & \multicolumn{2}{c|}{EFML} & \multicolumn{2}{c}{L-ELMo} \\
      Language & Cos & Euc & Cos & Euc \\
      \hline
        Croatian   & 0.652 & 73.54 & \textbf{0.428} & \textbf{33.48} \\
        English    & 0.442 & \textbf{23.89} & \textbf{0.432} & 42.99 \\
        Estonian   & 0.599 & 101.90 & \textbf{0.435} & \textbf{42.38}  \\
        Finnish    & 0.459 & \textbf{31.04} & \textbf{0.410} & 41.65  \\
        Latvian    & 0.494 & \textbf{30.32} & \textbf{0.466} & 42.75  \\
        Lithuanian & -     & -     & \textbf{0.389} & \textbf{29.37} \\
        Russian    & 0.495 & \textbf{29.47} & \textbf{0.429} & 44.24 \\
        Slovenian  & 0.568 & 99.22 & \textbf{0.408} & \textbf{28.16} \\
        Swedish    & 0.496 & \textbf{28.71} & \textbf{0.478} & 39.71 \\
      \hline
\end{tabular}
 \end{center}
\end{table}

The results for BERT models are presented in \Cref{tab:results-bert-monolingual-analogy}. Recall that the task for BERT models is different from ELMo, as described in \Cref{sec:evalSettingsAnalogies}. Each of the listed BERT models was used as a masked word prediction model in languages where that makes sense: monolingual and trilingual models were used for languages used in their pretraining, and massively multilingual models were tested for all used languages. BERTić (Croatian monolingual) model was not trained as a masked language model, so we omit it here.

\begin{table}[!htb]
\begin{center}
 \caption{The results of the word analogy task expressed as Accuracy@5 for multilingual BERT (mBERT), XLM-RoBERTa (XLM-R), trilingual BERT-based (TRI) and monolingual BERT-based (MONO) models. The best results for each language are typeset in \textbf{bold}.
 }
\label{tab:results-bert-monolingual-analogy}
\begin{tabular}{lrrrr}
      Language & mBERT & XLM-R & TRI & MONO \\
      \hline
      Croatian & 0.090 & 0.138 & \textbf{0.278} & - \\
      English & 0.404 & 0.413 & \textbf{0.439} & 0.114 \\ 
      Estonian & 0.093 & 0.251 & 0.224 & \textbf{0.393} \\ 
      Finnish & 0.067 & 0.208 & \textbf{0.285} & 0.173 \\
      Latvian & 0.026 & 0.118 & \textbf{0.170} & 0.118 \\
      Lithuanian & 0.036 & 0.107 & \textbf{0.214} & - \\
      Russian & 0.102 & \textbf{0.189} & - & 0.000 \\
      Slovenian & 0.061 & 0.146 & 0.195 & \textbf{0.409} \\
      Swedish & 0.052 & 0.097 & - & \textbf{0.239} \\
      \hline
\end{tabular}
\end{center}
\end{table}

The results show that trilingual BERT models are strongly dominating in most languages where they exist. The exceptions are Slovenian and Estonian, where the monolingual models perform best. However, in most languages monolingual models perform poorly. The latter holds also for EstBERT, which scores 0.165 accuracy@5, placing it behind every model, except mBERT. FinEst BERT is the best performing trilingual model on English for this task.

\subsubsection{SuperGLUE tasks}
\label{sec:results-superglue}
The SuperGLUE benchmark is extensively used to compare large pretrained models in English\footnote{\url{https://super.gluebenchmark.com/leaderboard}}. In contrast to that, we concentrate on the Slovene translation of the SuperGLUE tasks, described in \Cref{SuperGLUEbenchmark}. Experiments in English have shown that ELMo embeddings are not competitive to pretrained transformer models like BERT in GLUE benchmarks \citep{wang2019superglue}. For this reason, we skip ELMo models and compare four BERT models in our experiments: monolingual Slovene SloBERTa, trilingual CroSloEngual BERT, massively multilingual mBERT (bert-base-multilingual-cased\footnote{https://huggingface.co/bert-base-multilingual-cased}) and XLM-R  (xlm-roberta-base\footnote{https://huggingface.co/xlm-roberta-base}). Each model was fine-tuned using either MT or HT datasets of the same size. Only the translated content varies between both translation types; otherwise, they contain exactly the same examples. The splits of instances into train, validation and test sets is the same as in the English variant  (but mostly considerably smaller, see Table \ref{tab:superglue_translation}). 

In our analysis, we vary the sizes of datasets, translation types, and prediction models. Table \ref{tab:superglue_results} shows the results together with several baselines trained on the original English datasets. Most comparisons to English baselines are unfair because the reported English models used significantly more examples (BoolQ, MultiRC) or, in the case of the BERT++ model, the English model was additionally pretrained with transfer tasks that are similar to a target one (CB, RTE, BoolQ, COPA). In terms of comparable  datasets, fair comparisons are possible with the CB, COPA, and WSC.  

\begin{table}[htb]
\caption{The SuperGLUE benchmarks in English (upper part) and Slovene (lower part). All English results are taken from \citep{wang2019superglue}. The HT and MT labels indicate human and machine translated Slovene datasets. The best score for each task and language is in \textbf{bold}. The best average scores (Avg) for each language are \underline{underlined}. }
\label{tab:superglue_results}
\resizebox{\textwidth}{!}{%
\begin{tabular}{l|lllllllll}
\textbf{Task} & \textbf{Avg} & \textbf{BoolQ} & \textbf{CB} & \textbf{COPA} & \textbf{MultiRC} & \textbf{ReCoRD} & \textbf{RTE} & \textbf{WiC} & \textbf{WSC} \\
\textbf{Models/Metrics} &  & \textbf{Acc.} & \textbf{F1/Acc.} & \textbf{Acc.} & \textbf{F1$_a$/EM} & \textbf{F1/EM} & \textbf{Acc.} & \textbf{Acc.} & \textbf{Acc.} \\ \hline
Most Frequent  & 45.7 & 62.3 & 21.7/48.4 & 50.0 & 61.1/0.3 & 33.4/32.5 & 50.3 & 50.0 & \textbf{65.1} \\
CBoW  & 44.7 & 62.1 & 49.0/71.2 & 51.6 & 0.0/0.4 & 14.0/13.6 & 49.7 & 53.0 & \textbf{65.1} \\
BERT  & 69.3 & 77.4 & 75.7/83.6 & 70.6 & 70.0/24.0 & \textbf{72.0/71.3} & 71.6 & \textbf{69.5} & 64.3 \\
BERT++  & \underline{73.3} & \textbf{79.0} & \textbf{84.7/90.4} & \textbf{73.8} & \textbf{70.0/24.1} & \textbf{72.0/71.3} & \textbf{79.0} & \textbf{69.5} & 64.3 \\ Human (est.)  & 89.8 & 89.0 & 95.8/98.9 & 100.0 & 81.8*/51.9* & 91.7/91.3 & 93.6 & 80.0 & 100.0 \\ \hline 

Most Frequent (sl) &  49.1 & 63.3 & 21.7/48.4 & 50.0 & \textbf{76.4/0.6} & - & \textbf{58.6} & - & 65.8  \\ 
HT-mBERT                &  54.3 & 63.3 & 66.6/73.6 & 54.2 & 45.1/8.1 & - & 57.2 & - & 61.6  \\ 
MT-mBERT                &  55.2 & 63.3 & 65.1/68.8 & 54.4 & 55.4/11.7 & - & 57.9 & -  & -  \\ 
HT-CroSloEngual         &  55.6 & 63.3 & 62.1/72.4 & 58.2 & 53.0/8.4 & - &\textbf{58.6} & - & 56.2 \\ 
MT-CroSloEngual         &  53.4 & 63.3 & 59.8/68.4 & 55.0 & 51.2/10.5 & - & 53.8 & -  & -  \\ 
HT-SloBERTa             &  \underline{57.2} & 63.3 & \textbf{74.0/76.8} & \textbf{61.8} & 53.0/10.8 & - & 53.8 & - & \textbf{73.3} \\ 
MT-SloBERTa             &  55.8 & 63.3 & 68.6/74.8 & 58.2 & 57.1/12.0 & - & 49.6 & - & - \\ 
HT-XLM-R          &  53.5 & 63.3 & 66.2/73.2 & 50.0 & 53.3/0.9 & - & 57.2 & - & 65.8 \\ 
MT-XLM-R          &  50.1 & 63.3 & 62.0/68.4 & 51.4 & 55.3/0.6 & - & 42.8 & - & - \\ \hline

HT-Avg & \underline{55.1} & \textbf{63.3} & \textbf{70.6} & \textbf{56.0} & 29.1 & - & \textbf{56.7} & - & \textbf{64.2} \\ 
MT-Avg & 53.6 & \textbf{63.3} & 67.0 & 54.8 & \textbf{31.7} & - & 51.0 & - & - \\ \hline

\end{tabular}%
}
\end{table}

The single-number overall average score (Avg in the second column) comprises five equally weighted tasks: BoolQ, CB, COPA, MultiRC, and RTE. In tasks with multiple metrics, we averaged those metrics to get a single task score. For the details on how the score is calculated for each task, see \citep{wang2019superglue}.

Considering the Avg scores in Table \ref{tab:superglue_results}, SloBERTa is the best performing model. On average, all BERT models, regardless of translation type, perform better than the Most Frequent baselines. From the translation type perspective, the models trained on HT datasets perform better than those trained on MT datasets by 1.5 points.  
The only task where MT is better than HT is MultiRC, but looking at single scores, we can observe that none of the models learned anything (all scores are below the Most Frequent baseline). There is a large gap between the Most frequent baseline and the rest of the models. 
Analysis of other specific tasks shows that for the BoolQ dataset all models predict the most frequent class (the testing set might be too small for reliable conclusions in BoolQ). We assume that training sample sizes are too small MultiRC and BoolQ in these two tasks and have to be increased (we have only 92 HT examples in BoolQ and 15 HT examples in MultiRC). The same is also true for RTE. 

Compared to English models, the best Slovene model (SloBERTa) achieved good results on WSC. It seems that none of the English models learned anything from WSC, but the SloBERTa model achieved a score of 73.3 (the Most Frequent baseline gives 65.8). Nevertheless, there is still a large gap compared to human performance. All models showed good performance on CB and fell somewhere between English CBoW and BERT. We expected better results on the fully human translated COPA task. We are investigating the reasons for low performance in this task. In general, SloBERTa was significantly better than the rest of models in CB, COPA, and WSC. 

We conclude that the best BERT models perform well on tasks with enough training examples (CB, COPA, WSC) and show some level of language understanding above chance. Furthermore, the models benefited from human translated datasets compared to machine translation. For some datasets, we need to increase the number of training and/or testing examples. In further work, we intend to create a Slovene version of the WiC task from scratch and run experiments on the cleaned and improved ReCoRD task.

\subsection{Cross-lingual evaluations}
\label{crossEvaluation}
The cross-lingual evaluation is split into six subsections according to the type of task. We present results on NER,  POS-tagging, dependency parsing, analogies, SuperGLUE, and terminology alignment. We train models on a source language dataset in each task and use it for classification in the target language, i.e. we test the zero-shot transfer unless specified otherwise. For NER, POS-tagging, and dependency parsing tasks, the results are averaged over five individual evaluation runs, just like in the monolingual evaluations. Additionally, cross-lingual ELMoGAN maps were trained five times and each of the five maps was paired with one of the five classification models for each task during evaluation.

\subsubsection{Named entity recognition}
In \Cref{tab:results-elmo-xlingual-ner}, we present the results of cross-lingual transfer of contextual ELMo embeddings. We compared isomorphic mapping with Vecmap and MUSE libraries, and two non-isomorphic mappings using GANs, ELMoGAN-O (EG-O) and ELMoGAN-10k (EG-10k).

\begin{table}[h!tb]
\caption{Comparison of different methods for cross-lingual mapping of contextual ELMo embeddings evaluated on the NER task. The best Macro $F_1$ score for each language pair is in \textbf{bold}. The ``Reference`` column represents direct learning on the target language without cross-lingual transfer. The upper part of the table contains a scenario of cross-lingual transfer from English to a less-resourced language, and the lower part of the table shows a transfer between similar languages.} 
\centering
\resizebox{\linewidth}{!}{
\begin{tabular}{lllccccc}
Source & Target & Dict. & Vecmap & EG-O & EG-10k & MUSE & Reference \\
\hline
English & Croatian & direct & $\mathbf{0.385}$ & 0.274 & 0.365 & 0.024 & 0.810 \\
English & Estonian & direct & 0.554 & 0.693 & $\mathbf{0.728}$ & 0.284 & 0.895 \\
English & Finnish & direct & 0.672 & 0.705 & $\mathbf{0.780}$ & 0.229 & 0.922 \\
English & Latvian & direct & 0.499 & 0.644 & \textbf{0.652} & 0.216 & 0.818 \\
English & Lithuanian & direct & 0.498 & 0.522 & $\mathbf{0.553}$ & 0.208 & 0.755 \\
English & Slovenian & direct & 0.548 & 0.572 & $\mathbf{0.676}$ & 0.060 & 0.850 \\
English & Swedish & direct & \textbf{0.786} & 0.700 & 0.780 & 0.568 & 0.852 \\
\hline
Croatian & Slovenian & direct & 0.387 & 0.279 & 0.250 & \textbf{0.418} & 0.850 \\
Croatian & Slovenian & triang & $\mathbf{0.731}$ & 0.365 & 0.420 & 0.592 & 0.850 \\
Estonian & Finnish & direct & $\mathbf{0.517}$ & 0.339 & 0.316 & 0.278 & 0.922 \\
Estonian & Finnish & triang & $\mathbf{0.779}$ & 0.365 & 0.388 & 0.296 & 0.922 \\
Finnish & Estonian & direct & 0.477 & 0.305 & 0.324 & \textbf{0.506} & 0.895 \\
Finnish & Estonian & triang & \textbf{0.581} & 0.334 & 0.376 & 0.549 & 0.895 \\
Latvian & Lithuanian & direct & $\mathbf{0.423}$ & 0.398 & 0.404 & 0.345 & 0.755 \\
Latvian & Lithuanian & triang & \textbf{0.569} & 0.445 & 0.472 & 0.378 & 0.755 \\
Lithuanian & Latvian & direct & 0.263 & 0.312 & 0.335 & \textbf{0.604} & 0.818 \\
Lithuanian & Latvian & triang & 0.359 & 0.405 & 0.409 & \textbf{0.710} & 0.818 \\
Slovenian & Croatian & direct & 0.361 & 0.270 & 0.307 & \textbf{0.485} & 0.810 \\
Slovenian & Croatian & triang & $\mathbf{0.566}$ & 0.302 & 0.321 & 0.518 & 0.810 \\
\hline
\multicolumn{7}{l}{Average gap for the best cross-lingual transfer in each language} & 0.147 \\
\end{tabular}
}
\label{tab:results-elmo-xlingual-ner}
\end{table}

The upper part of the table shows a typical cross-lingual transfer learning scenario, where the model is transferred from resource-rich language (English) to less-resourced languages. In this case, the non-isomorphic ELMoGAN methods, particularly the ELMoGAN-10k variant, are superior to isomorphic mapping with Vecmap and MUSE libraries. In this scenario, ELMoGAN-10k is always the best or close to the best mapping approach. This is not always the case in the lower part of \Cref{tab:results-elmo-xlingual-ner}, which shows the second most important cross-lingual transfer scenario: transfer between similar languages. In this scenario, isomorphic mappings with Vecmap and MUSE are superior. We hypothesise that the reason for the better performance of isomorphic mappings is the similarity of tested language pairs and less violation of the isomorphism assumption the Vecmap and MUSE methods make. The results of the mapping with the MUSE method support this hypothesis. While MUSE performs worst in most cases of transfer from English, the performance gap is smaller for transfer between similar languages. MUSE is sometimes the best method for similar languages, but its results fluctuate considerably between language pairs. The second possible factor explaining the results is the quality of the dictionaries, which are in general better for combinations involving English. In particular, dictionaries obtained by triangulation via English are of poor quality, and non-isomorphic translation might be more affected by imprecise anchor points.

In general, even the best cross-lingual ELMo models lag behind the reference model without cross-lingual transfer. The differences in Macro $F_1$ score are small for some languages (e.g., 5.5\% for English-Swedish), but they are significantly larger for most languages. The average gap between the best cross-lingual model in each language and the monolingual reference is 14.7\% for ELMo models.

In \Cref{tab:results-bert-xlingual-ner}, we present the results of cross-lingual transfer for contextual BERT models. We compared massively multilingual BERT models (mBERT and XLM-R) with trilingual BERT models (TRI). 

\begin{table}[h!tb]
\caption{Comparison of multilingual BERT (mBERT), XLM-RoBERTa (XLM-R) and trilingual BERT-based (TRI) models evaluated on the NER task as a zero-shot transfer mode. The best Macro $F_1$ score for each language pair is in \textbf{bold}. The ``Best monolingual`` column represents the best result for direct learning on the target language without cross-lingual transfer. The upper part of the table contains a scenario of cross-lingual transfer from English to a less-resourced language, and the lower part of the table shows a transfer between similar languages.} 
\centering
\begin{tabular}{llccc|c}
 & & & & & Best \\
Source. & Target. & mBERT & XLM-R & TRI & monolingual \\
\hline
English & Croatian   & 0.632 & 0.673 & \textbf{0.814} & 0.886\\
English & Estonian   & 0.799 & \textbf{0.833} & 0.832 & 0.936\\
English & Finnish  & 0.780 & 0.840 & \textbf{0.902} & 0.957\\
English & Latvian   & 0.714 & 0.756 & \textbf{0.768} & 0.863 \\
English & Lithuanian  & 0.672 & 0.656 & \textbf{0.702} & 0.863\\
English & Slovenian  & 0.742 & 0.755 & \textbf{0.847} & 0.933 \\
\hline
Slovenian & Croatian & 0.751 & 0.769 & \textbf{0.841} & 0.886\\
Finnish & Estonian & 0.809 & 0.833 & \textbf{0.869} & 0.936\\
Estonian & Finnish  & 0.832 & 0.881 & \textbf{0.911} & 0.957\\
Lithuanian & Latvian & 0.785 & 0.816 & \textbf{0.834} & 0.863\\
Latvian & Lithuanian & 0.718 & 0.731 & \textbf{0.776} & 0.863\\
Croatian & Slovenian  & 0.844 & 0.882 & \textbf{0.901} & 0.933\\
\hline
\multicolumn{5}{l|}{Average gap for the best cross-lingual transfer in each language} & 0.052\\
\end{tabular}
\label{tab:results-bert-xlingual-ner}
\end{table}

The results show a clear advantage of trilingual models compared to massively multilingual models. The trilingual models dominate in 11 out of 12 transfers, except the transfer from English to Estonian, where XLM-R is better for 0.1\%. The results also show that the transfer from a similar language is more successful than transfer from English. The average difference between the most successful transfer from English and the most successful transfer from a similar language averaged over target languages is considerable, i.e. 4.6\%.

Comparing cross-lingual transfer of ELMo (in \Cref{tab:results-elmo-xlingual-ner}) with variants of multilingual BERT (in \Cref{tab:results-bert-xlingual-ner}), the transfer with BERT is considerably more successful. This indicates that ELMo, while useful for explicit extraction of embedding vectors, is less competitive with BERT in the prediction model transfer, especially if we consider that ELMo requires additional effort for preparation of contextual mapping datasets, while BERT does not need it.

Finally, the comparison between the best cross-lingual models (in the bottom part of \Cref{tab:results-bert-xlingual-ner}) and the best monolingual models (reference scores taken from \Cref{tab:results-bert-monolingual-ner}) shows that with cross-lingual transfer we lose on average 5.2\%. This is a very encouraging result, showing that modern cross-lingual technologies have made significant progress and can bridge the technological gap for less-resourced languages. Further, this score is for zero-shot transfer, while a few-shot transfer (with small amounts of data in a target language) might be even closer to monolingual results.

\subsubsection{POS-tagging}
In \Cref{tab:results-elmo-xlingual-pos}, we present the results of cross-lingual transfer of contextual ELMo embeddings. We compared isomorphic mapping with Vecmap and MUSE libraries and two non-isomorphic mappings using GANs (ELMoGAN-O and ELMoGAN-10k), described in \Cref{sec:ELMo}. The upper part of the table shows a  cross-lingual transfer learning scenario, where the model is transferred from resource-rich language (English) to less-resourced languages, and the lower part shows the transfer from similar languages.

\begin{table}[h!tb]
\caption{Comparison of different methods for cross-lingual mapping of contextual ELMo embeddings evaluated on the POS-tagging task. The best micro $F_1$ score for each language pair is in \textbf{bold}. The ``Reference`` column represents direct learning on the target language without cross-lingual transfer. The upper part of the table contains a scenario of cross-lingual transfer from English to a less-resourced language, and the lower part of the table shows a transfer between similar languages.} 
\centering
\resizebox{0.9\linewidth}{!}{
\begin{tabular}{lllccccc}
Source & Target & Dict. & Vecmap & EG-O & EG-10k & MUSE & Reference \\
\hline
English & Croatian & direct   & \textbf{0.705} & 0.629 & 0.620 & 0.687 & 0.963 \\
English & Estonian & direct   & 0.728 & 0.678 & 0.647 & \textbf{0.729} & 0.969 \\
English & Finnish & direct    & \textbf{0.729} & 0.531 & 0.578 & 0.715 & 0.966 \\
English & Latvian & direct    & \textbf{0.681} & 0.625 & 0.607 & 0.655 & 0.940 \\
English & Lithuanian & direct & \textbf{0.700} & 0.648 & 0.605 & 0.670 & 0.316 \\
English & Russian & direct    & 0.415 & 0.488 & 0.491 & \textbf{0.665} & 0.929 \\
English & Slovenian & direct  & 0.719 & 0.637 & 0.584 & \textbf{0.723} & 0.966 \\
English & Swedish & direct    & 0.839 & 0.688 & 0.649 & \textbf{0.848} & 0.933 \\
\hline
Croatian & Slovenian & direct & 0.551 & 0.421 & 0.435 & \textbf{0.683} & 0.966 \\
Croatian & Slovenian & triang & 0.734 & 0.434 & 0.461 & \textbf{0.833} & 0.966 \\
Estonian & Finnish & direct   & 0.586 & 0.522 & 0.533 & \textbf{0.706} & 0.966 \\
Estonian & Finnish & triang   & 0.673 & 0.514 & 0.543 & \textbf{0.690} & 0.966 \\
Finnish & Estonian & direct   & 0.619 & 0.596 & 0.590 & \textbf{0.792} & 0.969 \\
Finnish & Estonian & triang   & 0.703 & 0.603 & 0.583 & \textbf{0.837} & 0.969 \\
Latvian & Lithuanian & direct & 0.620 & 0.668 & 0.630 & \textbf{0.757} & 0.316 \\
Latvian & Lithuanian & triang & 0.657 & 0.671 & 0.646 & \textbf{0.773} & 0.316 \\
Lithuanian & Latvian & direct & 0.310 & 0.312 & 0.315 & \textbf{0.321} & 0.940 \\
Lithuanian & Latvian & triang & 0.303 & 0.316 & 0.315 & \textbf{0.319} & 0.940 \\
Slovenian & Croatian & direct & 0.558 & 0.467 & 0.495 & \textbf{0.662} & 0.963 \\
Slovenian & Croatian & triang & 0.735 & 0.492 & 0.502 & \textbf{0.784} & 0.963 \\
\hline
\multicolumn{7}{l}{Avg. gap for the best transfer in each language} & 0.104 \\
\multicolumn{7}{l}{Avg. gap for the best transfer in each language (without Lithuanian)} & 0.184 \\
\end{tabular}
}
\label{tab:results-elmo-xlingual-pos}
\end{table}

The isomorphic mappings with MUSE are superior in the POS tagging task, followed by Vecmap. The non-isomorphic methods are inferior in this task. However, even the best cross-lingual ELMo models lag considerably compared to the reference model without cross-lingual transfer. The average difference in Macro $F_1$ score is 22.3\% (not taking into account Lithuanian which has a failed monolingual ELMo model). 

In \Cref{tab:results-bert-xlingual-pos}, we present the results of cross-lingual transfer for contextual BERT models. We compared massively multilingual BERT models (mBERT and XLM-R) with trilingual BERT models (TRI).

\begin{table}[h!tb]
\caption{Comparison of multilingual BERT (mBERT), XLM-RoBERTa (XLM-R) and trilingual BERT-based (TRI) models evaluated on the POS-tagging task as a zero-shot knowledge transfer. The best $F_1$ score for each language pair is in \textbf{bold}. The upper part of the table contains a scenario of cross-lingual transfer from English to a less-resourced language, and the lower part of the table shows a transfer between similar languages.} 
\centering
\begin{tabular}{llccc|c}
& & \multicolumn{2}{c}{} & & Best \\
Source. & Target. & mBERT & XLM-R & TRI & monolingual \\
\hline
English & Croatian   & 0.837 & \textbf{0.846} & 0.827 & 0.982 \\
English & Estonian   & 0.799 & 0.849 & \textbf{0.851} & 0.977 \\
English & Finnish  & 0.799 & \textbf{0.857} & 0.839 & 0.980 \\
English & Latvian   & 0.756 & 0.828 & \textbf{0.829} & 0.966 \\
English & Lithuanian  & 0.766 & \textbf{0.833} & 0.822 & 0.964 \\
English & Russian & 0.812 & \textbf{0.842} & - & 0.976 \\
English & Slovenian  & 0.807 & \textbf{0.834} & 0.819 & 0.991 \\
English & Swedish & 0.908 & \textbf{0.925} & - & 0.981 \\
\hline
Slovenian & Croatian & 0.900 & 0.910 & \textbf{0.921} & 0.982 \\
Finnish & Estonian & 0.834 & 0.887 & \textbf{0.898} & 0.977 \\
Estonian & Finnish  & 0.813 & 0.889 & \textbf{0.890} & 0.980 \\
Lithuanian & Latvian & 0.805 & 0.857 & \textbf{0.860} & 0.966 \\
Latvian & Lithuanian & 0.821 & 0.890 & \textbf{0.900} & 0.964 \\
Croatian & Slovenian  & 0.895 & 0.919 & \textbf{0.924} & 0.991 \\
\hline
\multicolumn{5}{l|}{Average gap for the best cross-lingual transfer in each language} & 0.082 \\
\end{tabular}
\label{tab:results-bert-xlingual-pos}
\end{table}

The results show an advantage of trilingual models in transfer from similar languages, while in the transfer from English, the massively multilingual XLM-R models are more successful.  The transfer from a similar language is more successful than the transfer from English, the average difference being 5.7\%. 

Similarly to NER, the comparison of ELMo cross-lingual transfer (in \Cref{tab:results-elmo-xlingual-pos}) with variants of multilingual BERT (in \Cref{tab:results-bert-xlingual-pos}) shows that the transfer with BERT is considerably more successful. 
The comparison between the best cross-lingual models (these are various BERT models in \Cref{tab:results-bert-xlingual-pos}) and the best monolingual models (reference scores taken from \Cref{tab:results-bert-monolingual-pos}) shows that with the cross-lingual transfer we lose on average 8.2\%.

\subsubsection{Dependency parsing}
In \Cref{tab:results-elmo-xlingual-dp}, we present the results of cross-lingual transfer of contextual ELMo embeddings. We compared isomorphic mapping with Vecmap and MUSE libraries and two non-isomorphic mappings using GANs (ELMoGAN-O and ELMoGAN-10k). The upper part of the table shows a cross-lingual transfer learning scenario, where the model is transferred from resource-rich language (English) to less-resourced languages, and the lower part shows the transfer from similar languages.

\begin{table}[htbp]
\caption{Comparison of different contextual cross-lingual mapping methods for contextual ELMo embeddings, evaluated on the dependency parsing task. Results are reported as the unlabelled attachments score (UAS) and labelled attachment score (LAS). The best results for each language and type of transfer (from English or similar language) are typeset in \textbf{bold}. The column  ``Direct`` stands for direct learning on the target (i.e. evaluation) language without cross-lingual transfer. The languages are represented with their \href{https://en.wikipedia.org/wiki/List_of_ISO_639-1_codes}{international language codes ISO 639-1}.}
\centering
\resizebox{\linewidth}{!}{
\begin{tabular}{llc|rr|rr|rr|rr|rr}
Train & Eval. &  & \multicolumn{2}{c}{Vecmap} & \multicolumn{2}{c}{EG-O} & \multicolumn{2}{c}{EG-10k} & \multicolumn{2}{c}{MUSE} & \multicolumn{2}{c}{Direct}\\
lang. & lang. & Dict. & UAS & LAS & UAS & LAS & UAS & LAS & UAS & LAS & UAS & LAS \\ \hline
en & hr & direct & \textbf{73.96} & \textbf{60.53} & 68.73 & 50.29 & 66.74 & 40.93 & 71.01 & 54.89 & 91.74 & 85.84 \\
en & et & direct & \textbf{62.08} & \textbf{40.62} & 52.01 & 30.22 & 44.80 & 24.59 & 58.76 & 34.07 & 89.54 & 85.45 \\
en & fi & direct & \textbf{64.40} & \textbf{45.32} & 50.80 & 25.23 & 42.65 & 22.66 & 55.03 & 37.61 & 90.83 & 86.86 \\
en & lv & direct & \textbf{77.84} & \textbf{65.97} & 68.51 & 49.47 & 67.09 & 39.41 & 76.26 & 63.45 & 88.85 & 82.82 \\
en & lt & direct & \textbf{63.33} & \textbf{40.56} & 50.04 & 31.26 & 50.04 & 31.26 & 58.70 & 37.78 & 82.84 & 72.16 \\
en & ru & direct & \textbf{72.00} & \textbf{16.62} & 60.74 & 8.92 & 60.68 & 8.18 & 65.23 & 14.77 & 89.33 & 83.54 \\
en & sl & direct & \textbf{79.01} & \textbf{59.84} & 68.82 & 48.20 & 67.04 & 43.34 & 77.18 & 56.53 & 93.70 & 91.39 \\
en & sv & direct & 82.08 & 72.74 & 74.39 & 59.70 & 73.81 & 59.63 & \textbf{82.17} & \textbf{72.78} & 89.70 & 85.07 \\ \hline
hr & sl & direct & \textbf{85.47} & \textbf{72.70} & 51.88 & 31.50 & 53.68 & 33.40 & 83.45 & 69.08 & 93.70 & 91.39 \\
hr & sl & triang & \textbf{87.70} & \textbf{76.51} & 54.34 & 36.32 & 59.61 & 38.83 & \textbf{87.70} & 76.40 & 93.70 & 91.39 \\
et & fi & direct & \textbf{79.14} & \textbf{66.09} & 55.67 & 36.85 & 51.35 & 30.66 & 76.66 & 60.01 & 90.83 & 86.86 \\
et & fi & triang & \textbf{80.94} & \textbf{67.35} & 52.63 & 29.94 & 52.83 & 28.70 & 76.96 & 63.37 & 90.83 & 86.86 \\
fi & et & direct & \textbf{75.81} & 57.32 & 54.69 & 33.99 & 53.27 & 32.28 & 74.96 & \textbf{58.14} & 89.54 & 85.45 \\
fi & et & triang & \textbf{79.04} & \textbf{61.86} & 53.64 & 32.73 & 53.86 & 30.13 & 76.74 & 60.27 & 89.54 & 85.45 \\
lv & lt & direct & \textbf{76.43} & \textbf{54.24} & 64.44 & 37.16 & 64.73 & 35.86 & 75.45 & 53.02 & 82.84 & 72.16 \\
lv & lt & triang & \textbf{76.26} & \textbf{53.59} & 65.91 & 37.91 & 65.45 & 33.62 & 75.12 & 51.14 & 82.84 & 72.16 \\
lt & lv & direct & 63.27 & 24.53 & 56.43 & 26.93 & 62.51 & 31.84 & \textbf{73.70} & \textbf{44.62} & 88.85 & 82.82 \\
lt & lv & triang & 61.32 & 27.29 & 61.89 & 29.39 & 61.95 & 30.11 & \textbf{72.39} & \textbf{43.15} & 88.85 & 82.82 \\
sl & hr & direct & \textbf{77.89} & \textbf{62.58} & 47.34 & 29.39 & 52.27 & 32.48 & 72.87 & 55.70 & 91.74 & 85.84 \\
sl & hr & triang & \textbf{81.32} & \textbf{67.51} & 50.96 & 32.82 & 56.17 & 35.96 & 78.63 & 63.96 & 91.74 & 85.84 \\
\hline
\multicolumn{11}{l|}{Avg. gap for the best cross-lingual transfer in each language} & 9.89 & 23.79 \\
\end{tabular}
}
\label{tab:results-elmo-xlingual-dp}
\end{table}

The isomorphic mappings with Vecmap are superior in the dependency parsing task, followed by MUSE. Similarly to POS-tagging, the non-isomorphic methods lag. Again, the best cross-lingual ELMo models produce considerably lower scores than the reference model without cross-lingual transfer. The average difference in UAS score is 9.89\%, and in LAS it is 23.79\%. 

In \Cref{tab:results-bert-xlingual-dp1}, we present the results of cross-lingual transfer for contextual BERT models. We compared massively multilingual BERT models (mBERT and XLM-R) with trilingual BERT models (TRI).

\begin{table}[h!tbp]
\caption{Comparison of multilingual BERT (mBERT), XLM-RoBERTa (XLM-R) and trilingual BERT-based (TRI) models evaluated on the dependency parsing task as a zero-shot knowledge transfer. The best LAS score for each language pair is in \textbf{bold}. The upper part of the table contains a scenario of cross-lingual transfer from English to a less-resourced language, and the lower part of the table shows a transfer between similar languages.} 
\centering
\begin{tabularx}{\textwidth}{ll*{3}{Y}|c}
& & & & & Best \\
Source. & Target. & mBERT & XLM-R & TRI & monolingual \\
\hline
English & Croatian   & 42.13 & 54.00 & \textbf{56.04} & 82.36 \\
English & Estonian   & 25.12 & 38.01 & \textbf{42.30} & 78.64 \\
English & Finnish    & 29.08 & 43.30 & \textbf{46.18} & 83.64 \\
English & Latvian    & 23.06 & 38.66 & \textbf{44.93} & 74.32 \\
English & Lithuanian & 23.21 & 35.98 & \textbf{40.92} & 61.66 \\
English & Russian    & 43.41 & \textbf{48.19} & - & 80.90 \\
English & Slovenian  & 38.72 & 53.90 & \textbf{58.02} & 85.38 \\
English & Swedish    & 60.96 & \textbf{70.79} & - & 80.93 \\
\hline
Slovenian & Croatian & 52.61 & 63.66 & \textbf{67.60} & 82.36 \\
Finnish & Estonian & 37.34 & 53.98 & \textbf{63.08} & 78.64 \\
Estonian & Finnish  & 42.11 & 59.54 & \textbf{67.91} & 83.64 \\
Lithuanian & Latvian & 31.26 & 48.10 & \textbf{52.33} & 74.32 \\
Latvian & Lithuanian & 28.28 & 48.59 & \textbf{54.17} & 61.66 \\
Croatian & Slovenian  & 52.33 & 67.16 & \textbf{71.76} & 85.38 \\
\hline
\multicolumn{5}{l|}{Avg. gap for the best cross-lingual transfer in each language} & 16.50 \\
\end{tabularx}
\label{tab:results-bert-xlingual-dp1}
\end{table}

The results show an advantage of trilingual models in transfer from English and similar languages.  The transfer from a similar language is more successful than the transfer from English, the average difference being 14.74\%. The comparison between the best BERT cross-lingual models (from \Cref{tab:results-bert-xlingual-dp1}) and the best monolingual models (reference scores taken from \Cref{tab:results-bert-monolingual-dp1}) shows that with the cross-lingual transfer we lose on average 16.50\%.

Contrary to other tasks and similarly to monolingual setting, the comparison of ELMo cross-lingual transfer (in \Cref{tab:results-elmo-xlingual-dp}) with variants of multilingual BERT (in \Cref{tab:results-bert-xlingual-dp1}) shows that the transfer with ELMo is more successful. We hypothesise that this is the result of better ELMo source language models.

\subsubsection{Cross-lingual analogies}

We present the results of cross-lingual transfer of contextual ELMo embeddings in \Cref{tab:results-elmo-xlingual-analogy}. We compared isomorphic mapping with Vecmap and MUSE libraries and two non-isomorphic mappings using GANs (ELMoGAN-O and ELMoGAN-10k). The upper part of the table shows a cross-lingual transfer between English and lower-resourced language. The lower part of the table shows a cross-lingual transfer between two similar languages.

\begin{table}[htb]
\caption{Comparison of different contextual cross-lingual mapping methods for contextual ELMo embeddings, evaluated on the cross-lingual analogy task. Results are reported as the macro average distance between expected and actual word vector of the word w4. Two distance metrics were used: cosine (Cos) and Euclidean (Euc). The best results (shortest distance) for each language and type of transfer (from English or similar language) are typeset in \textbf{bold}. The column  ``Direct`` stands for monolingual evaluation on the target (i.e. evaluation) language without cross-lingual transfer. The languages are represented with their \href{https://en.wikipedia.org/wiki/List_of_ISO_639-1_codes}{international language codes ISO 639-1}.}
\centering
\resizebox{\linewidth}{!}{
\begin{tabular}{llc|rr|rr|rr|rr|rr}
Train & Eval. &  & \multicolumn{2}{c}{Vecmap} & \multicolumn{2}{c}{EG-O} & \multicolumn{2}{c}{EG-10k} & \multicolumn{2}{c}{MUSE} & \multicolumn{2}{c}{Direct}\\
lang. & lang. & Dict. & Cos & Euc & Cos & Euc & Cos & Euc & Cos & Euc & Cos & Euc \\ \hline
en & hr & direct & \textbf{0.603} & \textbf{23.47} & 0.814 & 40.02 & 0.763 & 42.40 & \textbf{0.603} & 44.54 & 0.428 & 33.48 \\
en & et & direct & \textbf{0.578} & \textbf{27.44} & 0.791 & 43.74 & 0.752 & 45.01 & 0.588 & 51.32 & 0.435 & 42.38 \\
en & fi & direct & 0.645 & 59.26 & 0.745 & \textbf{39.21} & 0.694 & 40.82 & \textbf{0.588} & 52.45 & 0.410 & 41.65 \\
en & lv & direct & 0.635 & \textbf{21.46} & 0.809 & 44.62 & 0.778 & 46.58 & \textbf{0.623} & 50.79 & 0.466 & 42.75 \\
en & lt & direct & 0.697 & \textbf{30.39} & 0.812 & 38.67 & 0.719 & 40.84 & \textbf{0.598} & 41.55 & 0.389 & 29.37 \\
en & ru & direct & \textbf{0.573} & 64.35 & 0.771 & \textbf{41.49} & 0.705 & 43.28 & 0.574 & 53.20 & 0.429 & 44.24 \\
en & sl & direct & \textbf{0.613} & \textbf{32.29} & 0.836 & 38.42 & 0.731 & 40.07 & 0.664 & 42.92 & 0.408 & 28.16 \\
en & sv & direct & 0.615 & 64.66 & 0.787 & \textbf{37.35} & 0.720 & 38.84 & \textbf{0.587} & 47.11 & 0.478 & 39.71 \\
\hline
hr & sl & direct & 0.690 & \textbf{7.59} & 0.732 & 41.02 & 0.721 & 41.29 & \textbf{0.592} & 36.37 & 0.408 & 28.16 \\
hr & sl & triang & 0.715 & \textbf{23.89} & 0.729 & 40.91 & 0.727 & 41.45 & \textbf{0.564} & 35.22 & 0.408 & 28.16 \\
et & fi & direct & \textbf{0.545} & \textbf{11.08} & 0.796 & 47.04 & 0.775 & 48.08 & 0.549 & 50.27 & 0.410 & 41.65 \\
et & fi & triang & 0.816 & \textbf{33.33} & 0.799 & 46.50 & 0.759 & 47.97 & \textbf{0.527} & 49.06 & 0.410 & 41.65 \\
fi & et & direct & 0.598 & \textbf{11.27} & 0.685 & 41.99 & 0.653 & 43.10 & \textbf{0.551} & 48.47 & 0.435 & 42.38 \\
fi & et & triang & 0.692 & \textbf{30.25} & 0.725 & 41.23 & 0.644 & 43.09 & \textbf{0.554} & 48.42 & 0.435 & 42.38 \\
lv & lt & direct & 0.587 & \textbf{11.96} & 0.704 & 39.52 & 0.624 & 41.80 & \textbf{0.563} & 39.99 & 0.389 & 29.37 \\
lv & lt & triang & 0.681 & \textbf{19.77} & 0.711 & 39.81 & 0.621 & 41.77 & \textbf{0.570} & 40.17 & 0.389 & 29.37 \\
lt & lv & direct & 0.690 & \textbf{12.10} & 0.814 & 45.38 & 0.758 & 46.86 & \textbf{0.524} & 43.26 & 0.466 & 42.75 \\
lt & lv & triang & 0.704 & \textbf{18.18} & 0.812 & 45.28 & 0.752 & 46.47 & \textbf{0.525} & 43.36 & 0.466 & 42.75 \\
sl & hr & direct & 0.591 & \textbf{6.62} & 0.663 & 38.00 & 0.645 & 38.23 & \textbf{0.526} & 38.17 & 0.428 & 33.48 \\
sl & hr & triang & 0.572 & \textbf{20.02} & 0.665 & 37.45 & 0.651 & 38.44 & \textbf{0.501} & 36.92 & 0.428 & 33.48 \\
\hline
\multicolumn{11}{l|}{Average gap for the best cross-lingual transfer in each language} & 0.118 & -20.29 \\
\end{tabular}
}
\label{tab:results-elmo-xlingual-analogy}
\end{table}

The results depend largely on the metric used for evaluation. With cosine distance, the mappings with MUSE are the best in most cases. For language pairs, where the MUSE method is not the best, it is a close second. However, with Euclidean distance, Vecmap mappings perform the best in most language pairs, especially between similar languages, where they significantly outperform even monolingual results. This can be partially explained by the fact, that the Vecmap method changes both the source and target language embeddings during the mapping. For three language pairs, English-Finnish, English-Russian, and English-Swedish, Vecmap mappings do not perform well using the Euclidean distance. In those cases, ELMoGAN-O mappings perform best.

In \Cref{tab:results-bert-xlingual-analogy}, we present the results of contextual BERT models on the cross-lingual analogy task. We compared massively multilingual BERT models (mBERT and XLM-R) with  trilingual BERT models: Croatian-Slovene-English (CSE), Finnish-Estonian-English (FinEst), and Lithuanian-Latvian-English (LitLat). 
Recall that in the cross-lingual setting, the word analogy task tries to match each relation in one language with each relation from the same category in the other language. For cross-lingual contextual mappings, the word analogy task is less adequate, and we apply this task to words in invented contexts. The upper part of the table shows a cross-lingual scenario from the resource-rich language (English) to less-resourced languages, and the lower part shows the transfer from similar languages.

\begin{table}[htb]
\caption{Comparison of multilingual BERT (mBERT), XLM-RoBERTa (XLM-R) and trilingual BERT-based (TRI) models, evaluated on the word analogy task as a zero-shot knowledge transfer. The best accuracy@5 score for each language pair is in \textbf{bold}. The upper part of the table contains a scenario of cross-lingual transfer from English to a less-resourced language, and the lower part of the table shows a transfer between similar languages.} 
\centering
\begin{tabular}{llccc|c}
& & \multicolumn{2}{c}{} & & Best \\
Source. & Target. & mBERT & XLM-R & TRI & monolingual \\
\hline
English & Croatian   & 0.025 & 0.015 & \textbf{0.103} & 0.278 \\
English & Estonian   & 0.018 & 0.029 & \textbf{0.074} & 0.393 \\
English & Finnish    & 0.001 & 0.013 & \textbf{0.114} & 0.285 \\
English & Latvian    & 0.006 & \textbf{0.036} & 0.033 & 0.170 \\
English & Lithuanian & 0.011 & 0.034 & \textbf{0.042} & 0.214 \\
English & Russian    & 0.045 & \textbf{0.088} & - & 0.189 \\
English & Slovenian  & 0.007 & 0.055 & \textbf{0.091} & 0.409 \\
English & Swedish    & \textbf{0.065} & 0.053 & - & 0.239 \\
\hline
Slovenian & Croatian  & 0.024 & 0.088 & \textbf{0.139} & 0.278 \\
Finnish & Estonian    & 0.019 & 0.035 & \textbf{0.073} & 0.393 \\
Estonian & Finnish    & 0.003 & 0.020 & \textbf{0.137} & 0.285 \\
Lithuanian & Latvian  & 0.005 & 0.016 & \textbf{0.032} & 0.170 \\
Latvian & Lithuanian  & 0.011 & 0.033 & \textbf{0.068} & 0.214 \\
Croatian & Slovenian  & 0.013 & 0.086 & \textbf{0.178} & 0.409 \\
\hline
\multicolumn{5}{l}{Average gap for the best cross-lingual transfer in each language} & 0.174 \\
\end{tabular}
\label{tab:results-bert-xlingual-analogy}
\end{table}

The results show an advantage of trilingual models in transfer from both English and similar languages (the only difference being the transfer from English to Latvian, where the XLM-R is more successful).  The transfer from a similar language is mostly more successful than the transfer from English.

\subsubsection{SuperGLUE tasks}
In the cross-lingual scenario, we tested three models, mBERT, CroSloEngual BERT (CSE), and XLM-R, in the model transfer between English and Slovene (both directions). 
For Slovene as the source language, we used the available human translated examples. To make the comparison balanced, we only used the same examples from English datasets. We tested both zero-shot transfer (no training data in the target language) and few-shot transfer. In the few-shot training, we used 10 additional examples from the target language for each task. To achieve more statistically valid results, we randomly sampled these 10 examples five times and reported averages. The fine-tuning hyperparameters are the same as in the monolingual setup. 

The results are presented in Table \ref{tab:cross-lingual_superglue}. Averaged over all tasks, some models improved the Most frequent baseline. In general, they were quite unsuccessful on BoolQ, MultiRC, and WSC but showed some promising results on COPA, RTE, and especially CB. Additional training examples in the few-shot scenario brought some visible improvements. It seems that models perform better in the English-Slovene direction than vice versa. The best performing model is XLM-R, followed by CroSloEngual BERT and mBERT. 

\begin{table}[htb]
\centering
\caption{Cross-lingual results on human translated SuperGLUE test sets. The source (s.) and target (t.) language is either Slovene (sl) or English (en). The best results for zero-shot and few-shot scenarios are in bold. }
\label{tab:cross-lingual_superglue}
\resizebox{\textwidth}{!}{%
\begin{tabular}{llll|lllllll}
\multirow{2}{*}{\textbf{Transfer}} & \multirow{2}{*}{\textbf{Model}} & \multirow{2}{*}{\textbf{s.}} & \multirow{2}{*}{\textbf{t.}} & \textbf{Avg} & \textbf{BoolQ} & \textbf{CB} & \textbf{COPA} & \textbf{MultiRC} & \textbf{RTE} & \textbf{WSC} \\
 &  &  &  &  & \textbf{acc.} & \textbf{F1/acc.} & \textbf{Acc.} & \textbf{F1$_a$/EM} & \textbf{Acc.} & \textbf{Acc.} \\ \hline
\multirow{6}{*}{Zero-shot} & 

 \multirow{2}{*}{CSE} & en & sl & 49.8 & 56.7 & 43.7/60.0 & 54.6 & 48.0/6.6 & 58.6 & 50.7 \\
 &  & sl & en & 52.6 & 60.0 & 53.8/70 & \textbf{59.6} & \textbf{56.7/9.6} & 48.3 & 58.2 \\
 
 & \multirow{2}{*}{mBERT} & en & sl & 47.4 & 56.7 & 36.2/57.2 & 50.2 & 47.3/8.7 & 55.2 & 64.4 \\
 &  & sl & en & 48.3 & 60.0 & 44.6/50.4 & 49.8 & 56.2/8.7 & 51.7 & 57.5 \\ 
 
  & \multirow{2}{*}{XLM-R} & en & sl & \textbf{53.8} & \textbf{63.3} & \textbf{62.9/68.4} & 53.6 & 48.5/0.3 & \textbf{62.1} & 56.2 \\
 & & sl & en & 51.7 & \textbf{63.3} & 59.1/67.2 & 47.2 & 52.9/12.9 & 51.7 & \textbf{65.8} \\ \hline
 
\multirow{6}{*}{Few-shot} & 

\multirow{2}{*}{CSE} & en & sl & 54.4 & 60.0 & 52.4/68.6 & 55.0 & 52.8/9.72 & \textbf{65.5} & 54.1 \\
 &  & sl & en & 53.0& 60.0	& 53.8/70.0 &	\textbf{59.5} &	56.0/12.1 &	49.7 &   58.2 \\

 & \multirow{2}{*}{mBERT} & en & sl & 50.9 & 60.1 & 53.1/66.2 & 50.4 & 50.8/9.8 & 53.8 & 64.4 \\
 &  & sl & en & 51.3 & 60.7 &	51.8/58.2 &	50.3 &	57.2/11.1&	56.5&	56.8
 \\
 
  & \multirow{2}{*}{XLM-R} & en & sl & \textbf{57.0} & \textbf{63.3} & \textbf{65.8/69.8} & 53.3 & \textbf{76.4/0.6} & 62.1 & 57.4 \\
 &  & sl & en & 53.0 & \textbf{63.3} &	63.0/69.6&	48.3 &	51.4/10.6	& 55.8 &	\textbf{65.8}
 \\ \hline
 
 & \multicolumn{2}{l}{Most frequent}  &  & 52.4 & 63.3 & 23.0/52.7 & 50.0 & 77.3/0.3 & 58.6 & \textbf{65.8}
\end{tabular}
}
\end{table}

The low overall performance can be explained by a low number of training examples in the source language. If we take a closer look at specific models, we can observe that XLM-R shows good results on CB in both directions and evaluation scenarios. CroSloEngual BERT achieved similarly good result on COPA, where it is the only model that overpassed the baseline.     

We can conclude that for the difficult SuperGLUE benchmark, the cross-lingual transfer is challenging but not impossible. In the future, we plan to expand the current set of experiments in several directions. First, we will train English models on the full SuperGLUE datasets and transfer them to Slovene human and machine-translated datasets. Second, we will train Slovene models on the combined machine and human translated datasets and transfer them to full English datasets. We will combine Slovene and English training sets and apply the models to both languages. Finally, we will also combine training for several tasks and test transfer learning scenarios.

\subsubsection{Terminology alignment}

We present the results of cross-lingual terminology alignment of contextual ELMo embeddings in \Cref{tab:results-elmo-xlingual-terminology}. We compared the same four mapping methods as in the previous tasks. For each language pair, we evaluate the terminology alignment in both directions. That is, given the terms from the first language (source), we search for the equivalent terms in the second language (target), then we repeat in the other direction.

\begin{table}[!!h!tb]
\caption{Comparison of contextual cross-lingual mapping methods for ELMo embeddings, evaluated on the terminology alignment task. Results are reported as accuracy@1, based on the cosine distance metric. The best results for each language and type of transfer (from English in the upper part or from a similar language in the lower part) are typeset in \textbf{bold}. The languages are represented with their \href{https://en.wikipedia.org/wiki/List_of_ISO_639-1_codes}{international language codes ISO 639-1}. Direction in the third column represents the direction of vector mapping: from$\rightarrow$to.}
\centering
\resizebox{0.8\linewidth}{!}{
\begin{tabular}{llc|cccc}
Source & Target & Dictionary & Vecmap & EG-O & EG-10k & MUSE \\
lang. & lang. & (direction) &  &  &  & \\ \hline
en & sl & direct (sl$\rightarrow$en) & 0.079 & \textbf{0.152} & 0.151 & 0.096 \\
sl & en & direct (sl$\rightarrow$en) & 0.099 & 0.139 & \textbf{0.195} & 0.126 \\
en & hr & direct (hr$\rightarrow$en) & 0.080 & \textbf{0.153} & 0.135 & 0.116 \\
hr & en & direct (hr$\rightarrow$en) & 0.084 & 0.139 & \textbf{0.153} & 0.102 \\
en & et & direct (et$\rightarrow$en) & 0.092 & \textbf{0.177} & 0.167 & 0.128 \\
et & en & direct (et$\rightarrow$en) & 0.091 & 0.117 & \textbf{0.133} & 0.118 \\
en & fi & direct (fi$\rightarrow$en) & 0.092 & 0.166 & \textbf{0.176} & 0.132 \\
fi & en & direct (fi$\rightarrow$en) & 0.087 & 0.083 & \textbf{0.116} & 0.112 \\
en & lv & direct (lv$\rightarrow$en) & 0.084 & \textbf{0.157} & 0.147 & 0.102 \\
lv & en & direct (lv$\rightarrow$en) & 0.091 & 0.122 & \textbf{0.140} & 0.111 \\
en & lt & direct (lt$\rightarrow$en) & 0.095 & \textbf{0.181} & 0.172 & 0.114 \\
lt & en & direct (lt$\rightarrow$en) & 0.097 & 0.132 & \textbf{0.171} & 0.102 \\
en & sv & direct (sv$\rightarrow$en) & 0.125 & 0.183 & \textbf{0.187} & 0.161 \\
sv & en & direct (sv$\rightarrow$en) & 0.112 & 0.111 & \textbf{0.167} & 0.109 \\
\hline
sl & hr & direct (hr$\rightarrow$sl) & \textbf{0.109} & 0.037 & 0.031 & 0.102 \\
sl & hr & triang (hr$\rightarrow$sl) & 0.130 & 0.056 & 0.046 & \textbf{0.156} \\
sl & hr & direct (sl$\rightarrow$hr) & \textbf{0.109} & 0.039 & 0.038 & 0.100 \\
sl & hr & triang (sl$\rightarrow$hr) & 0.130 & 0.053 & 0.057 & \textbf{0.155} \\
hr & sl & direct (hr$\rightarrow$sl) & \textbf{0.084} & 0.029 & 0.028 & 0.082 \\
hr & sl & triang (hr$\rightarrow$sl) & 0.097 & 0.042 & 0.044 & \textbf{0.121} \\
hr & sl & direct (sl$\rightarrow$hr) & \textbf{0.084} & 0.023 & 0.021 & \textbf{0.084} \\
hr & sl & triang (sl$\rightarrow$hr) & 0.097 & 0.039 & 0.033 & \textbf{0.121} \\
fi & et & direct (et$\rightarrow$fi) & \textbf{0.130} & 0.092 & 0.078 & 0.121 \\
fi & et & triang (et$\rightarrow$fi) & \textbf{0.130} & 0.102 & 0.080 & 0.124 \\
fi & et & direct (fi$\rightarrow$et) & \textbf{0.129} & 0.085 & 0.089 & 0.122 \\
fi & et & triang (fi$\rightarrow$et) & 0.130 & 0.090 & 0.094 & \textbf{0.145} \\
et & fi & direct (et$\rightarrow$fi) & 0.143 & 0.091 & 0.094 & \textbf{0.167} \\
et & fi & triang (et$\rightarrow$fi) & 0.148 & 0.095 & 0.103 & \textbf{0.166} \\
et & fi & direct (fi$\rightarrow$et) & 0.143 & 0.108 & 0.092 & \textbf{0.166} \\
et & fi & triang (fi$\rightarrow$et) & 0.148 & 0.118 & 0.097 & \textbf{0.189} \\
lv & lt & direct (lt$\rightarrow$lv) & 0.102 & 0.080 & 0.061 & \textbf{0.123} \\
lv & lt & triang (lt$\rightarrow$lv) & 0.119 & 0.090 & 0.076 & \textbf{0.134} \\
lv & lt & direct (lv$\rightarrow$lt) & 0.102 & 0.059 & 0.071 & \textbf{0.123} \\
lv & lt & triang (lv$\rightarrow$lt) & 0.119 & 0.065 & 0.077 & \textbf{0.128} \\
lt & lv & direct (lt$\rightarrow$lv) & 0.099 & 0.061 & 0.069 & \textbf{0.102} \\
lt & lv & triang (lt$\rightarrow$lv) & 0.112 & 0.064 & 0.076 & \textbf{0.116} \\
lt & lv & direct (lv$\rightarrow$lt) & 0.099 & 0.071 & 0.057 & \textbf{0.102} \\
lt & lv & triang (lv$\rightarrow$lt) & \textbf{0.112} & 0.083 & 0.069 & 0.110 \\

\hline
\end{tabular}
}
\label{tab:results-elmo-xlingual-terminology}
\end{table}

Results show that for the terminology alignment between English and other languages, the two non-isomorphic mappings perform the best on all language pairs. With English as the target language, ELMoGAN-10k always performs the best. In cases where English is the source language, ELMoGAN-O is usually the best.
For terminology alignment between similar languages, 
isomorphic methods outperform the non-isomorphic methods on similar languages. In most cases, MUSE is the best method. If we just look at the best dictionary and mapping direction for each language pair, MUSE is the best in each language pair not involving English.

The terminology alignment is in most cases better from English than from a similar language as the source. The exceptions are Croatian and Finnish (as the targets).

In \Cref{tab:results-bert-xlingual-terminology}, we present the results of contextual embeddings, extracted from multilingual and trilingual BERT models. In the same table, we also compare the BERT embeddings with the best ELMo alignments. The results show that trilingual models significantly outperform massively multilingual models. The exception is alignment between Latvian (source) and Lithuanian (target), where mBERT and LitLat-BERT perform equally. Compared to ELMo embeddings, trilingual BERT models achieve better results on alignment between similar languages. However, ELMo outperforms BERT models on most language pairs, where source terms are in English (the exceptions are Croatian and Slovenian). The mBERT model performs poorly in most cases.

\begin{table}[htb]
\caption{Comparison of mBERT, XLM-R, and trilingual BERT-based models (TRI), evaluated on the terminology alignment task. Results are reported as accuracy@1, based on the cosine distance metric. The best results for each language pair are typeset in \textbf{bold}. The languages are represented with their \href{https://en.wikipedia.org/wiki/List_of_ISO_639-1_codes}{international language codes ISO 639-1}. The best ELMo result for each language pair (from \Cref{tab:results-elmo-xlingual-terminology}) is in the rightmost column. The best overall results for each language pair are \underline{underlined}.}
\centering
\begin{tabular}{ll|ccc|c}
Source lang. & Target lang. & mBERT & XLM-R & TRI & best ELMo \\ \hline
en & hr & 0.054 & 0.049 & \underline{\textbf{0.187}} & 0.153 \\
hr & en & 0.029 & 0.073 & \underline{\textbf{0.230}} & 0.153 \\
en & et & 0.057 & 0.064 & \textbf{0.121} & \underline{0.177} \\
et & en & 0.069 & 0.086 & \underline{\textbf{0.183}} & 0.133 \\
en & fi & 0.084 & 0.090 & \textbf{0.146} & \underline{0.176} \\
fi & en & 0.026 & 0.105 & \underline{\textbf{0.215}} & 0.116 \\
en & lv & 0.068 & 0.059 & \textbf{0.107} & \underline{0.157} \\
lv & en & 0.016 & 0.088 & \underline{\textbf{0.156}} & 0.140 \\
en & lt & 0.072 & 0.058 & \textbf{0.099} & \underline{0.181} \\
lt & en & 0.016 & 0.081 & \textbf{0.147} & \underline{0.171} \\
en & sl & 0.060 & 0.055 & \underline{\textbf{0.195}} & 0.152 \\
sl & en & 0.103 & 0.098 & \underline{\textbf{0.284}} & 0.195 \\
en & sv & 0.135 & \textbf{0.151} & - & \underline{0.187} \\
sv & en & 0.063 & \underline{\textbf{0.178}} & - & 0.167 \\ 
\hline
sl & hr & 0.251 & 0.143 & \underline{\textbf{0.267}} & 0.156 \\
hr & sl & 0.099 & 0.124 & \underline{\textbf{0.250}} & 0.121 \\
fi & et & 0.063 & 0.130 & \underline{\textbf{0.217}} & 0.145 \\
et & fi & 0.150 & 0.145 & \underline{\textbf{0.233}} & 0.189 \\
lt & lv & 0.177 & 0.128 & \underline{\textbf{0.206}} & 0.116 \\
lv & lt & \underline{\textbf{0.195}} & 0.133 & \underline{\textbf{0.195}} & 0.134 \\
\hline
\end{tabular}
\label{tab:results-bert-xlingual-terminology}
\end{table}

\section{Conclusions}
\label{sec:conclusions}

We performed a large scale evaluation of monolingual and cross-lingual contextual embedding approaches on several languages with sufficient resources. We concentrated on recently most successful contextual embeddings, in particular ELMo and BERT models. For ELMo models, we compared monolingual embeddings from two sources and several cross-lingual mappings with and without assumption of isomorphism. For BERT models, we compared monolingual models, massively multilingual models, and trilingual models. In the evaluation, we used several tasks: NER, POS-tagging, dependency parsing, CoSimLex, analogies, terminology alignment, and the SuperGLUE benchmarks. 

Overall, the results show that L-ELMo models are superior to other ELMo models, but in general, there is a clear advantage of BERT models over ELMo models. In the monolingual setting, monolingual and trilingual BERT models are very competitive, and frequently the trilingual BERT models dominate. In the cross-lingual setting, BERT models are much more successful compared to ELMo models. The trilingual models are mostly better than the massively multilingual models. There are a few exceptions to these general conclusions. The main outlier is the dependency parsing task, where the L-ELMo embeddings are better than BERT models. 
The results also indicate that for the training of BERT-like models, large enough datasets are a prerequisite. For example, 500 million tokens used for LVBERT is not enough and even 1.1 billion tokens used for EstBERT does not guarantee good performance.

We can conclude that cross-lingual transfer of trained prediction models is feasible with the presented approaches, especially from similar languages and using specifically designed trilingual models. For several tasks, the performance of the best cross-lingual transferred models lags behind the monolingual models by only a few percent, confirming findings described in \citep{robnik2021slovenscina20}. The exact performance drop depends on the task and language.

In future work, it would be worth testing other forms of cross-lingual transfer, in particular different degrees of few-shot learning. While we compare the cross-lingual transfer of models with the human translation baselines in SuperGLUE tasks, a wider comparison using more tasks would be welcome.

\section*{Acknowledgments} 
The work was partially supported by the Slovenian Research Agency (ARRS) core research programmes P6-0411 and and P2-0103, as well as the Ministry of Culture of Republic of Slovenia through project Development of Slovene in Digital Environment (RSDO). Partial support was also received from the UK EPSRC under grant EP/S033564/1, and the EPSRC/AHRC Centre for Doctoral Training in Media and Arts Technology EP/L01632X/1.
This paper is supported by European Union's Horizon 2020 research and  innovation programme under grant agreement No 825153, project EMBEDDIA (Cross-Lingual Embeddings for Less-Represented Languages in European News Media).
The results of this publication reflect only the authors' view and the EU Commission is not responsible for any use that may be made of the information it contains.

\bibliography{NLP,embeddia-refs}
\bibliographystyle{plainnat}

\end{document}